\documentclass[letterpaper]{article} 
\usepackage[]{aaai2026}  
\usepackage{times}  
\usepackage{helvet}  
\usepackage{courier}  
\usepackage[hyphens]{url}  
\usepackage{graphicx} 
\usepackage{natbib}  
\usepackage{caption} 
\usepackage{algorithm}

\usepackage{newfloat}
\usepackage{listings}
\DeclareCaptionStyle{ruled}{labelfont=normalfont,labelsep=colon,strut=off} 
\floatstyle{ruled}
\newfloat{listing}{tb}{lst}{}
\floatname{listing}{Listing}
\usepackage[utf8]{inputenc} 
\usepackage[T1]{fontenc}    

\usepackage{url}            
\usepackage{booktabs}       
\usepackage{amsfonts}       
\usepackage{nicefrac}       
\usepackage{microtype}      
\usepackage{xcolor}         
\usepackage{markdown}
\usepackage[colorinlistoftodos, loadshadowlibrary]{todonotes}
\usepackage{amsmath}
\usepackage{algorithm}
\usepackage{algpseudocode}

\presetkeys{todonotes}{size=\tiny}{}

\usepackage{graphicx}
\usepackage[most]{tcolorbox}
\definecolor{skyblue}{RGB}{95, 157, 241}
\usepackage{multirow}
\usepackage[inline]{enumitem}
\usepackage{algcompatible}
\usepackage{placeins}
\title{Can Hallucinations Help? Boosting LLMs for Drug Discovery}
\author{
    Shuzhou Yuan, Zhan Qu, Ashish Yashwanth Kangen, Michael Färber}
\affiliations{
    ScaDS.AI and TU Dresden\\

    \{shuzhou.yuan, zhan.qu, michael-faerber\}@tu-dresden.de
}

\usepackage{bibentry}

\begin{document}

\maketitle

\begin{abstract}
Hallucinations in large language models (LLMs), plausible but factually inaccurate text, are often viewed as undesirable. However, recent work suggests that such outputs may hold creative potential. In this paper, we investigate whether hallucinations can improve LLMs on molecule property prediction, a key task in early-stage drug discovery. We prompt LLMs to generate natural language descriptions from molecular SMILES strings and incorporate these often hallucinated descriptions into downstream classification tasks. Evaluating seven instruction-tuned LLMs across five datasets, we find that hallucinations significantly improve predictive accuracy for some models. Notably, \texttt{Falcon3-Mamba-7B} outperforms all baselines when hallucinated text is included, while hallucinations generated by \texttt{GPT-4o} consistently yield the greatest gains between models. We further identify and categorize over 18,000 beneficial hallucinations, with \textit{structural misdescriptions} emerging as the most impactful type, suggesting that hallucinated statements about molecular structure may increase model confidence. Ablation studies show that larger models benefit more from hallucinations, while temperature has a limited effect. Our findings challenge conventional views of hallucination as purely problematic and suggest new directions for leveraging hallucinations as a useful signal in scientific modeling tasks like drug discovery.

\end{abstract}


\section{Introduction}

\begin{figure}[ht]
    \centering
    \includegraphics[width=1\linewidth]{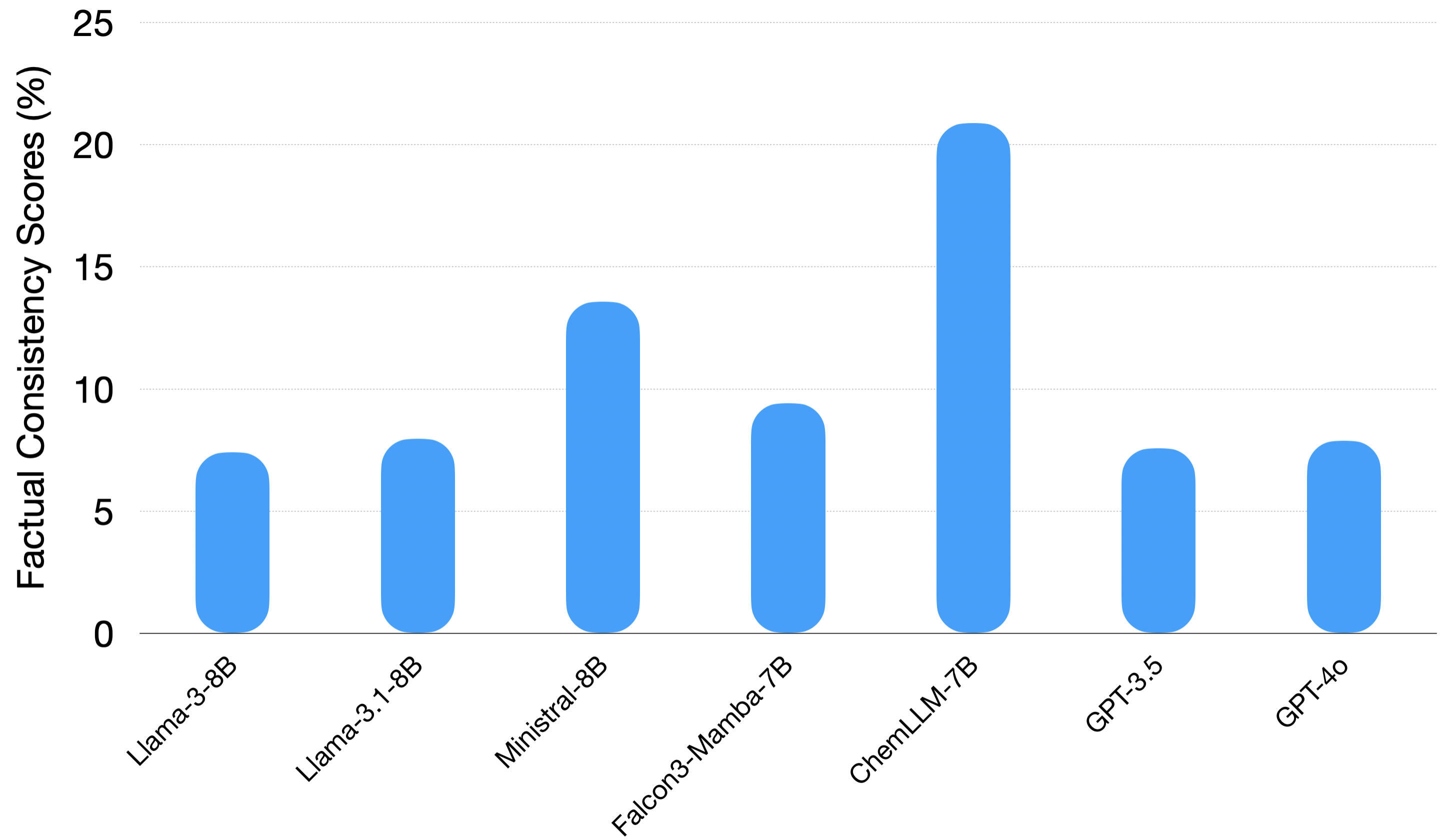}
    \caption{Factual consistency scores evaluated using the HHEM-2.1-Open Model, measuring the degree of hallucination relative to \texttt{MolT5} reference descriptions. Scores are averaged across five datasets. Except for \texttt{ChemLLM}, all LLMs exhibit low factual consistency, indicating extensive hallucination.}
    \label{figure:intro_figure}
\end{figure}



Large language models (LLMs) have demonstrated impressive capabilities across diverse domains, including everyday applications~\citep{NEURIPS2023_1190733f,yao2024survey}, scientific discovery~\citep{zhang-etal-2024-honeycomb,madani2023large}, and chemistry~\citep{boiko2023autonomous,zhang2024chemllm}. However, LLMs are known to produce hallucinations, text that is plausible but factually incorrect or misaligned with the input~\citep{huang2023survey,bang-etal-2023-multitask,rawte-etal-2023-troubling,yuan-faerber-2023-evaluating}. While much work has focused on mitigating hallucinations~\citep{ji-etal-2023-towards,dhuliawala-etal-2024-chain}, recent perspectives suggest that hallucination may also be beneficial, for example, it may enable creativity~\citep{lee2023mathematical,ye2023cognitive,wang2024lighthouse}.


In domains like drug discovery, creativity is a vital asset. Evaluating molecular properties for potential drug candidates involves reasoning over vast chemical spaces and abstract functional traits~\citep{hughes2011principles}. LLMs are beginning to assist in this process~\citep{chakraborty2023artificial,edwards-etal-2024-language}, and textual molecule descriptions have been shown to help improve generalization~\citep{liu2023multi}. Yet, many LLM-generated descriptions contain hallucinations, as illustrated in Figure~\ref{figure:intro_figure}, where models deviate significantly from domain-specific references such as \texttt{MolT5}~\citep{edwards-etal-2022-translation}.

This raises an underexplored question: \textit{Can hallucinations improve LLMs' performance in molecule property prediction?} While hallucinations are often treated as failure cases, we hypothesize that they may act as implicit counterfactuals, offering speculative, high-level interpretations that help LLMs generalize over unseen compounds. This idea is particularly relevant in early-stage drug discovery, where structured knowledge is sparse and creativity may be an advantage.

\begin{figure*}[ht]
    \centering
    \includegraphics[width=0.9\linewidth]{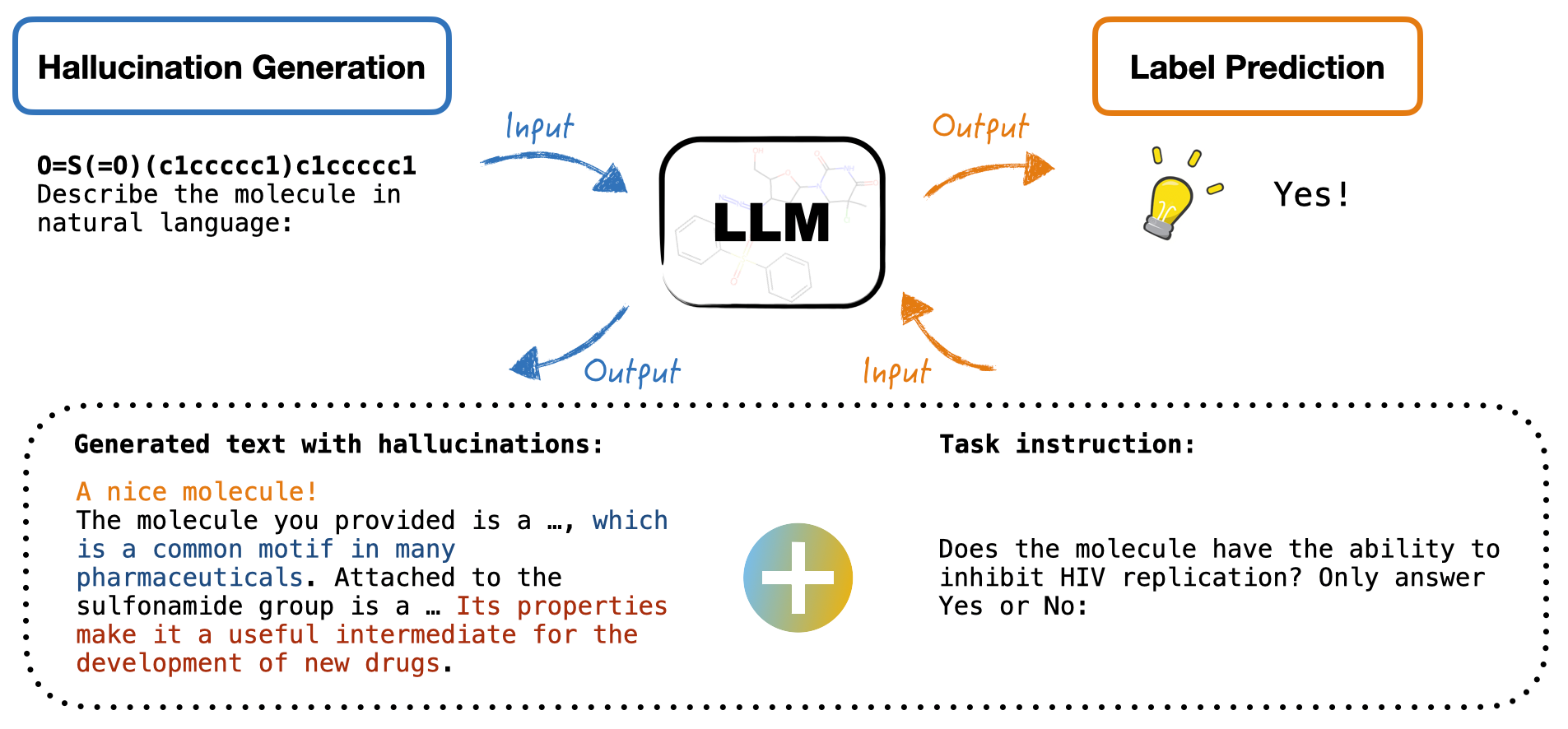}
    \caption{Illustration of our method using a sample from the HIV dataset. We first generate hallucinated molecule descriptions from LLMs, then include them as part of the prompt for a binary classification task. Output is constrained to ``Yes'' or ``No.'' Different types of hallucinations are color-coded.}
    \label{figure:methodology}
\end{figure*}

To investigate this, we conduct a comprehensive evaluation of seven instruction-tuned LLMs across five molecular property prediction tasks from MoleculeNet~\citep{wu2018moleculenet}. Given a molecule’s SMILES string~\citep{weininger1988smiles}, we generate natural language descriptions using each LLM, then include the description in the prompt alongside a task-specific instruction (Figure~\ref{figure:methodology}). We compare the resulting performance against three baselines:
\begin{itemize}
    \item \texttt{SMILES}: using only the molecule string,
    \item \texttt{MolT5}: domain-specific reference descriptions,
    \item \texttt{PubChem}: rule-based text derived from authoritative chemical metadata~\citep{kim2025pubchem}.
\end{itemize}

Our results show that LLM-generated hallucinations can enhance performance. For example, \texttt{Falcon3-Mamba-7B} surpasses the \texttt{PubChem} baseline by 8.22\% in ROC-AUC. \texttt{Llama-3.1-8B} achieves 15.8\% and 11.2\% improvements over the \texttt{SMILES} and \texttt{MolT5} baselines, respectively. Hallucinations from \texttt{GPT-4o} deliver the most consistent gains across models.

We further conduct ablation studies and analyze the effects of model size and generation temperature. Larger models benefit more from hallucinations, while temperature affects factuality (hallucination rate) but has limited influence on downstream performance.

To understand the nature of these helpful hallucinations, we collect instances that lead to improved predictions and categorize them into four types: \textit{structural misdescription}, \textit{functional hallucination}, \textit{analogical hallucination}, and \textit{generic fluff}. Surprisingly, the dominant type is \textit{structural misdescription}, suggesting that factual deviation is not necessarily detrimental, and may help LLMs generalize.

\vspace{0.5em}
\noindent\textbf{Our contributions are as follows:}
\begin{itemize}
  \item We conduct the first systematic investigation into whether hallucinations can improve LLMs for molecular property prediction, a core subtask in early-stage drug discovery.
  \item We evaluate seven LLMs across five benchmark datasets, demonstrating that certain models benefit significantly from hallucinated input.
  \item We analyze model size, generation temperature, and hallucination types to understand how and when hallucinations contribute positively to performance.
\end{itemize}

\section{Related Work}
\paragraph{LLMs for Molecules} 
Large language models have shown promising potential in molecular tasks and biomedical domains~\citep{pal2023chatgpt,murakumo2023llm,chakraborty2023artificial,savage2023drug}. \citet{zheng2024large} provide a comprehensive review of LLMs across the drug discovery pipeline, while \citet{guan2024drug} discuss their advantages in drug development. Several works evaluate LLMs on molecular property prediction~\citep{zhong2024benchmarking}, or apply LLMs to specialized tasks such as dataset curation~\citep{liu2024drugagent} and multi-agent drug discovery frameworks~\citep{liu2024drugagent}. Natural language explanations have also been explored for cancer drug discovery via prompt-based methods~\citep{zhang-etal-2024-prompt}. The connection between molecular representations such as SMILES strings and language has been widely investigated in recent works~\citep {edwards-etal-2022-translation,ganeeva-etal-2024-lost}, while LLMs have increasingly been used as general-purpose agents in the domain~\citep{fossi2024swiftdossier}. Building on this foundation, our work takes a novel step by examining how LLM-generated hallucinations can be leveraged to improve molecular property prediction. Rather than treating hallucinations as noise to eliminate, we explore their potential as creative and informative inputs in early-stage drug discovery.

\paragraph{Hallucination and Creativity in LLMs}
Hallucination is typically defined as a mismatch between the generated content and a source or ground truth~\citep{maynez-etal-2020-faithfulness,rawte-etal-2023-troubling}, and numerous approaches aim to detect or reduce hallucination~\citep{dhuliawala-etal-2024-chain,manakul-etal-2023-selfcheckgpt,ji-etal-2023-towards}. However, a growing body of work explores the creative potential of hallucinations~\citep{lee2022rethinking,zhao2024assessing}. \citet{jiang2024surveylargelanguagemodel} argue that hallucinations in LLMs parallel human creativity by enabling the recombination of known information in novel ways. Similarly, \citet{lee2023mathematical} model hallucination as a probabilistic driver of surprising or insightful outputs. Other studies evaluate LLMs’ creative capacity in narrative generation~\citep{chen-ding-2023-probing,gomez-rodriguez-williams-2023-confederacy,franceschelli2024creativity}. Our work builds on this perspective, exploring hallucinations not as factual errors, but as potentially productive deviations that improve downstream decision-making in scientific domains.

\paragraph{Counterfactual Reasoning and Hypothetical Text}
Counterfactual reasoning, which investigates how slight changes in input affect model predictions, has become a foundational method to explore causal relationships and to delineate decision boundaries in explainable AI~\citep{byrne2019counterfactuals,prado2024survey}. They offer actionable insights, reveal model biases, and can uncover failure cases~\citep{qucody,guidotti2024counterfactual}. 

While most applications of counterfactuals in NLP focus on fairness, robustness, or factual consistency~\citep{wu2021polyjuice,qiu2024paircfr,wang2024survey,chatzi2024counterfactual}, recent work begins to explore implicit counterfactuals in generative text~\citep{dixit2023counterfactual}. In molecular modeling, counterfactual explanations have been applied to explain property predictions by generating minimally perturbed molecular variants~\citep{wellawatte2022model,zhang2025mmgcf,teufel2025improving}. Hallucinated inputs, although not explicitly constructed as counterfactuals, may similarly function as soft perturbations that induce informative changes in model behavior. Hallucinated text can be interpreted as speculative hypotheses about molecular properties, enabling models to generalize beyond literal training examples. To our knowledge, our study is the first to examine this connection in the context of molecule property prediction with LLMs.

\section{Hallucination Generation}
\label{section:hallucination_generation}
Large language models (LLMs), when prompted to describe molecules, often produce fluent yet factually incorrect descriptions, commonly referred to as hallucinations. While typically regarded as undesirable, we explore an alternative view: these hallucinations can be seen as implicit counterfactuals, imaginative variations on molecular structure or function that may enhance model performance in downstream prediction tasks. In this section, we detail how we generate these descriptions and analyze their properties.

\subsection{Prompt Design}

Translating molecular structures into natural language has been shown to support various biomedical and scientific tasks~\citep{edwards-etal-2022-translation}. In this work, we follow a similar approach by prompting large language models (LLMs) to generate descriptive text from SMILES strings.


To ensure consistency across models and datasets, we use a standardized prompt template for all molecule descriptions. We also define the system role as an ``expert in drug discovery'' to encourage chemistry-relevant outputs. The prompt is shown below:

\begin{tcolorbox}[colback=skyblue!20, colframe=skyblue!80!black, width=0.48\textwidth, rounded corners]
\textit{System}: You are an expert in drug discovery.\\
\textit{User}: $[SMILES]$ Describe the molecule in natural language:
\end{tcolorbox}

We apply this prompt to generate molecule descriptions using seven different LLMs across five benchmark datasets. To assess the factual consistency of the generated text, we compare the outputs to reference descriptions generated by \texttt{MolT5}~\citep{edwards-etal-2022-translation}, a model pre-trained and fine-tuned on large-scale molecule-text pairs.



Factual consistency is measured using the HHM-2.1-Open model~\citep{hhem-2.1-open}, which estimates alignment between generated descriptions and reference text. As shown in Figure~\ref{figure:intro_figure}, most LLM-generated descriptions diverge substantially from the \texttt{MolT5} references. Even \texttt{ChemLLM}, which is domain-tuned, achieves only 20.89\% consistency. Other models range from 7.42\% to 13.58\%. These results confirm that the majority of generated descriptions contain some degree of hallucination.

Throughout this work, we define hallucination as any information in the generated text that is not supported by the molecule’s actual structure or known properties. We note that all descriptions are generated in English, as no language constraints were applied.

\subsection{Hallucination Typology and Annotation}

To better characterize the nature of hallucinated content, we define a typology of hallucination types based on qualitative analysis. We randomly sample 50 descriptions from across datasets and manually identify four recurring hallucination patterns. An additional fifth category, ``no hallucination,’’ is used to label fully accurate outputs. Table~\ref{table:hallucination_categories} summarizes the categories and provides representative examples.


\begin{table}[!hb]
\centering
\scalebox{0.6}{
\begin{tabular}{@{}lll@{}}
\toprule
\textbf{Type} & \textbf{Definition} \\ \midrule
Functional hallucination & Adds plausible usage/application info or unverified biochemical action \\
Structural misdescription & Incorrect description of atoms, bonds, or substructures \\
Analogical hallucination & Uses metaphor or analogy to describe structure or function \\
Generic fluff & Vague, uninformative, or subjective phrases \\
No hallucination & Accurate statements without speculative or incorrect elements \\ \bottomrule
\end{tabular}}
\caption{Categorization of hallucination types in molecular descriptions.}
\label{table:hallucination_categories}
\end{table}

We annotate hallucination types at scale using an external LLM, \texttt{Deepseek-R1}~\citep{guo2025deepseek}, which was not used to generate the descriptions. The annotation prompt includes the SMILES string, the generated description, and the type definitions in Table~\ref{table:hallucination_categories}. The model is instructed to assign one or more appropriate hallucination types to each input.

To assess annotation quality, we randomly sample 50 instances and have two human annotators independently assign hallucination types. Inter-annotator agreement with the model is measured using Fleiss’ Kappa, yielding $\kappa = 0.57$, which indicates moderate agreement and acceptable reliability.

\section{Task Formulation}
\label{section:task_formulation}

We define molecule property prediction as a prompt-based decision task for large language models (LLMs). Each instance consists of a molecule represented by a SMILES string $[SMILES]$, optionally augmented with a natural language description $[Description]$, followed by a task-specific instruction $[Instruct]$.

\begin{tcolorbox}[colback=skyblue!20, colframe=skyblue!80!black, width=0.48\textwidth, rounded corners]
\textit{System}: You are an expert in chemistry.\\
\textit{User}: $[SMILES]$\;$[Description]$\;$[Instruct]$
\end{tcolorbox}

The model is then asked to answer a binary yes/no question, such as: \textit{“Does this molecule inhibit HIV replication? Only answer `Yes' or `No'.”}

We decode the model's prediction by selecting the highest-probability token:

\begin{equation}
\label{argmax}
\hat{y} = \arg\max_{y \in V} P_\phi(y), \quad V = \{\text{Yes}, \text{No}\}
\end{equation}

\vspace{0.5em}
\noindent
\textbf{Textual Conditioning as Counterfactual Context.}
Our central hypothesis is that the presence of $[Description]$, regardless of factual accuracy, provides useful inductive bias for the model. These descriptions vary across three settings:

\begin{itemize}[leftmargin=1.5em]
    \item \textbf{None (SMILES-only)}: The model receives no textual context beyond the SMILES string.
    \item \textbf{Factual}: A structured, rule-based description from PubChem is included.
    \item \textbf{LLM-generated}: A free-text description is generated as described in the previous section, and may include hallucinations.
\end{itemize}

Rather than treating hallucinated descriptions as noise, we consider them as \textit{implicit counterfactuals}, alternative narratives about molecular properties that may help the model distinguish subtle patterns. This setup enables us to investigate how speculative or misaligned information interacts with LLM decision-making in molecular tasks.

\section{Experimental Design}
\label{section:exp_design}

We investigate how molecule-level hallucinations generated by different LLMs affect downstream classification performance. The prompt format is constructed as $[SMILES][Description][Instruct]$, where $[Description]$ is varied under four different settings:

\vspace{0.5em}
\noindent\textbf{\texttt{SMILES}}:
$[Description]$ is set to $\epsilon$ (empty), and the model must make predictions based solely on the SMILES string.

\vspace{0.5em}
\noindent\textbf{\texttt{MolT5}}:
We use descriptions generated by \texttt{MolT5}~\citep{edwards-etal-2022-translation}, a pretrained molecule-to-text model. This provides a reference point with high-quality factual descriptions aligned to molecular structure.

\vspace{0.5em}
\noindent\textbf{\texttt{PubChem}}:
We extract rule-based factual information from the PubChem database~\citep{kim2025pubchem}, including molecular formula, atom count, synonyms, and molecular weight. This metadata is formatted into a natural language description, e.g., \textit{“This compound, known as X, has the molecular formula C$_{22}$H$_{23}$N$_7$O$_2$…”}. While this setting reflects gold-standard chemical knowledge, it is limited to molecules present in PubChem (88\% coverage). As such, this baseline is not intended for practical deployment but to assess model behavior under ideal factual conditions.

\vspace{0.5em}
\noindent\textbf{\texttt{LLMs}}:
To study the effects of hallucinations, we replace $[Description]$ with text generated by different LLMs. Each model is evaluated using (a) its own hallucinated descriptions and (b) hallucinations generated by other LLMs. This setup allows us to analyze if performance is influenced more by the source of the hallucination or the model that interprets it.

\subsection{Datasets}
\label{section:dataset}

We select five benchmark datasets from MoleculeNet~\citep{wu2018moleculenet}, covering a range of physiological and pharmacological properties: \textbf{HIV}: Predicts anti-HIV activity based on inhibition assay data. \textbf{BBBP}: Indicates whether a molecule can penetrate the blood-brain barrier. \textbf{Clintox}: Labels compounds that failed clinical trials due to toxicity. \textbf{SIDER}: Classifies drug molecules by whether they cause specific side effects.\textbf{Tox21}: Measures toxicity across 12 biological targets.


\vspace{0.5em}
\noindent\textbf{Data Splits.}
We follow the recommended MoleculeNet protocol. HIV and BBBP are split using scaffold splitting~\citep{bemis1996properties}, which groups molecules based on core structure. Clintox, SIDER, and Tox21 use random splitting. For each dataset, we hold out 10\% of the data as the test set.

\vspace{0.5em}
\noindent\textbf{Label Selection and Evaluation.}
We retain the original binary labels for HIV, BBBP, and Clintox. Since SIDER and Tox21 contain multiple labels, we select the one with the most balanced distribution in the test set. For SIDER, this corresponds to the label \textit{“reproductive system and breast disorders”}. For Tox21, we use the \textit{SR-MMP} target (matrix metalloproteinases). We report model performance using ROC-AUC, as standard for imbalanced binary classification tasks~\citep{wu2018moleculenet}.

\subsection{Models}
\label{model_details}

We evaluate seven instruction-tuned LLMs, grouped by category: \textbf{General-purpose open-source LLMs}: \texttt{Llama-3-8B}, \texttt{Llama-3.1-8B}, \texttt{Mistral-8B}, and \texttt{Falcon3-Mamba-7B}. \textbf{Domain-specific LLM}: \texttt{ChemLLM-7B}, trained on chemical corpora. \textbf{Closed-source API models}: \texttt{GPT-3.5-turbo} and \texttt{GPT-4o}.


All open-source models are roughly 7--8B in size for fair comparison. Text generation is conducted using default decoding settings, with temperature 0.6 and a maximum token limit of 256. Models are evaluated both as \textit{generators} (to produce hallucinated descriptions) and as \textit{predictors} (to make property predictions given a prompt).

\section{Main Results and Analysis}\label{section:main_results}
We conduct experiments to address three primary questions: 

\begin{enumerate}[label={\textbf{\roman{*})}}]
    \item Can hallucinated descriptions improve LLM performance compared to standard baselines?
    \item Which LLMs generate the most beneficial hallucinations?
    \item What types of hallucinations are most likely to improve prediction?
\end{enumerate}

\begin{table*}[ht!]
\centering
\scalebox{0.8}{
\begin{tabular}{llccccccccc}
\toprule
\textbf{Model}                          & $\boldsymbol{[Description]}$ & \textbf{HIV}       & \textbf{BBBP}    & \textbf{Clintox} & \textbf{SIDER}   & \textbf{Tox21}   & \textbf{Avg}     & $\boldsymbol\Delta$\textbf{\texttt{SMILES}} & $\boldsymbol\Delta$\textbf{\texttt{MolT5}} & $\boldsymbol\Delta$\textbf{\texttt{PubChem}}   \\
\midrule
\multirow{6}{*}{\textbf{\texttt{Llama-3-8B}}} & \texttt{SMILES}    & \textbf{67.78} & 53.08          & \textbf{63.04} & \textbf{61.79} & 60.34          & \textbf{61.21} & -      & 6.53   & 0.55    \\
                                & \texttt{MolT5}     & 47.65          & 59.65          & 43.20          & \underline{61.14} & \underline{61.73} & 54.68          & -6.53  & -      & -5.98   \\
                                & \texttt{PubChem}   & 49.85          & \textbf{72.48} & 53.35          & 59.99          & \textbf{67.60} & \underline{60.65} & -0.55  & 5.98   & -       \\
                                & \texttt{Llama-3}   & \underline{53.13} & 58.21          & \underline{57.13} & 57.07          & 59.75          & 57.06          & -4.15  & 2.38   & -3.59   \\
                                & \texttt{ChemLLM}   & 46.83          & \underline{60.22} & 52.36          & 58.50          & 50.83          & 53.75          & -7.46  & -0.93  & -6.90   \\
                                & \texttt{GPT-4o}    & 45.97          & 53.72          & 54.81          & 45.86          & 60.90          & 52.25          & -8.96  & -2.42  & -8.40   \\
\midrule
\multirow{6}{*}{\textbf{\texttt{Llama-3.1-8B}}} & \texttt{SMILES}    & 38.10          & 37.56          & 35.30          & 52.70          & 44.89          & 41.71          & -      & -4.57  & -22.36  \\
                                & \texttt{MolT5}     & 43.04          & 45.30          & 39.28          & 52.06          & 51.72          & 46.28          & 4.57   & -      & -17.8   \\
                                & \texttt{PubChem}   & \textbf{58.94} & \textbf{71.42} & \underline{66.03} & \textbf{57.54} & \textbf{66.45} & \textbf{64.08} & 22.36  & 17.80  & -       \\
                                & \texttt{Llama-3}   & \underline{56.80} & 47.71          & \textbf{68.08} & \underline{54.21} & 60.75          & \underline{57.51} & 15.80  & 11.23  & -6.57   \\
                                & \texttt{ChemLLM}   & 50.83          & 52.18          & 47.45          & 53.17          & 49.84          & 50.69          & 8.98   & 4.41   & -13.39  \\
                                & \texttt{GPT-4o}    & 56.64          & \underline{53.37} & 55.28          & 52.60          & \underline{61.65} & 55.91          & 14.20  & 9.63   & -8.17   \\
\midrule
\multirow{6}{*}{\textbf{\texttt{Ministral-8B}}} & \texttt{SMILES}    & \textbf{59.35} & 48.87          & 60.19          & 57.27          & 57.42          & 56.62          & -      & 3.53   & -5.63   \\
                                & \texttt{MolT5}     & 44.54          & 44.10          & 53.95          & \underline{63.83} & 59.02          & 53.09          & -3.53  & -      & -9.16   \\
                                & \texttt{PubChem}   & 50.08          & \underline{52.77} & \textbf{79.23} & \textbf{64.45} & \textbf{64.72} & \textbf{62.25} & 5.63   & 9.16   & -       \\
                                & \texttt{Llama-3}   & \underline{52.91} & 31.27          & \underline{65.49} & 50.45          & 55.81          & 51.19          & -5.43  & -1.90  & -11.06  \\
                                & \texttt{ChemLLM}   & 46.35          & 48.66          & 62.11          & 58.97          & 55.36          & 54.29          & -2.33  & 1.20   & -7.96   \\
                                & \texttt{GPT-4o}    & 51.66          & \textbf{64.94} & 58.73          & 60.54          & \underline{61.07} & \underline{59.39} & 2.77   & 6.30   & -2.86   \\
\midrule
\multirow{6}{*}{\textbf{\texttt{Falcon3-Mamba-7B}}} & \texttt{SMILES}    & 40.64          & 48.33          & 31.72          & 52.53          & 47.45          & 44.13          & -      & 0.21   & -1.34   \\
                                & \texttt{MolT5}     & 49.02          & 47.92          & 18.58          & \underline{55.84} & 48.27          & 43.93          & -0.21  & -      & -1.54   \\
                                & \texttt{PubChem}   & 45.40          & \underline{58.84} & 24.55          & 52.00          & 46.56          & 45.47          & 1.34   & 1.54   & -       \\
                                & \texttt{Llama-3}   & 47.84          & 53.40          & \underline{50.76} & 51.31          & \underline{48.44} & 50.35          & 6.22   & 6.42   & 4.88    \\
                                & \texttt{ChemLLM}   & \textbf{53.46} & \textbf{61.82} & 45.12          & 53.56          & 43.43          & \underline{51.48} & 7.35   & 7.55   & 6.01    \\
                                & \texttt{GPT-4o}    & \underline{51.59} & 55.73          & \textbf{53.55} & \textbf{56.03} & \textbf{51.54} & \textbf{53.69} & 9.55   & 9.76   & 8.22    \\
\midrule
\multirow{6}{*}{\textbf{\texttt{ChemLLM-7B}}} & \texttt{SMILES}    & \underline{55.54} & 38.69          & 24.02          & 47.85          & \textbf{65.59} & 46.34          & -      & -1.93  & -8.36   \\
                                & \texttt{MolT5}     & 50.33          & 41.01          & 35.43          & \underline{53.06} & 61.50          & 48.27          & 1.93   & -      & -6.43   \\
                                & \texttt{PubChem}   & 54.96          & \textbf{72.27} & 34.37          & 51.08          & 60.79          & \textbf{54.69} & 8.36   & 6.43   & -       \\
                                & \texttt{Llama-3}   & \textbf{56.29} & 45.68          & \textbf{47.51} & 44.95          & 51.59          & 49.20          & 2.87   & 0.94   & -5.49   \\
                                & \texttt{ChemLLM}   & 45.37          & 41.83          & \underline{42.80} & \textbf{53.15} & 46.60          & 45.95          & -0.39  & -2.32  & -8.74   \\
                                & \texttt{GPT-4o}    & 52.68          & \underline{55.73} & 40.74          & 44.59          & \underline{62.63} & \underline{51.28} & 4.94   & 3.01   & -3.42   \\
\midrule
\multirow{6}{*}{\textbf{\texttt{GPT-3.5}}} & \texttt{SMILES}    & \textbf{61.32} & 28.94          & 63.90          & 53.77          & 61.92          & 53.97          & -      & -2.56  & -11.01  \\
                                & \texttt{MolT5}     & 49.34          & \underline{52.64} & \underline{64.30} & 56.00          & 60.39          & \underline{56.53} & 2.56   & -      & -8.44   \\
                                & \texttt{PubChem}   & \underline{56.64} & \textbf{61.11} & \textbf{76.34} & \underline{59.99} & \textbf{70.81} & \textbf{64.98} & 11.01  & 8.44   & -       \\
                                & \texttt{Llama-3}   & 52.55          & 30.00          & 38.27          & 52.22          & 55.71          & 45.75          & -8.22  & -10.78 & -19.23  \\
                                & \texttt{ChemLLM}   & 50.27          & 45.97          & 63.03          & 53.77          & 54.57          & 53.52          & -0.45  & -3.01  & -11.46  \\
                                & \texttt{GPT-4o}    & 49.67          & 34.61          & 60.83          & \textbf{62.41} & \underline{65.36} & 54.57          & 0.60   & -1.96  & -10.40  \\
\midrule
\multirow{6}{*}{\textbf{\texttt{GPT-4o}}} & \texttt{SMILES}    & 47.19          & 46.79          & 40.00          & 52.30          & 63.25          & 49.91          & -      & 2.10   & -6.17   \\
                                & \texttt{MolT5}     & 41.73          & 40.57          & 38.09          & \textbf{65.22} & 53.42          & 47.80          & -2.10  & -      & -8.27   \\
                                & \texttt{PubChem}   & 50.96          & 50.34          & \underline{56.67} & \underline{56.89} & \textbf{65.52} & \textbf{56.08} & 6.17   & 8.27   & -       \\
                                & \texttt{Llama-3}   & 48.95          & 36.23          & 52.97          & 42.73          & 62.00          & 48.58          & -1.33  & 0.77   & -7.5    \\
                                & \texttt{ChemLLM}   & \underline{53.00} & \textbf{54.92} & 55.59          & 45.36          & 61.15          & 54.00          & 4.10   & 6.20   & -2.08   \\
                                & \texttt{GPT-4o}    & \textbf{53.30} & \underline{52.73} & \textbf{57.12} & 47.82          & \underline{65.34} & \underline{55.26} & 5.35   & 7.46   & -0.82   \\
\bottomrule
\end{tabular}}
\caption{The main results (ROC-AUC \%) for a curated set of descriptions. For each model, we show the three baselines (\texttt{SMILES}, \texttt{MolT5}, \texttt{PubChem}) and three selected hallucinated descriptions (\texttt{Llama-3}, \texttt{ChemLLM}, \texttt{GPT-4o}). $\Delta$\texttt{SMILES}, $\Delta$\texttt{MolT5}, and $\Delta$\texttt{PubChem} denote the difference in average ROC-AUC scores compared to the baselines. Within each model group, the best and second-best scores for each dataset are highlighted in \textbf{bold} and \underline{underlined}, respectively.}
\label{tab:main_results}
\end{table*}

\FloatBarrier
\subsection{Do Hallucinations Improve LLMs?}
Table~\ref{tab:main_results} presents the performance of each model on five benchmark datasets, comparing hallucinated molecule descriptions against three baselines: the SMILES-only input, the neural reference \texttt{MolT5}, and the rule-based \texttt{PubChem} descriptions. To provide representative coverage across different model types, we include the LLM from each of the three categories whose generated hallucinations resulted in the greatest average performance gains in ROC-AUC: \texttt{Llama-3} for open-source models, \texttt{GPT-4o} for proprietary models, and \texttt{ChemLLM-7B} for domain-specific models fine-tuned on molecular data.\footnote{Full results for all hallucinated variants and model configurations are provided in the appendix.}

\texttt{PubChem} consistently serves as the strongest factual baseline, with 14 out of 35 model–dataset combinations (7 models across 5 datasets) achieving their highest performance when provided with its accurate, rule-based metadata. Nevertheless, hallucinated descriptions can occasionally surpass even this gold-standard input. For example, \texttt{Falcon3-Mamba} achieves an average ROC-AUC that is 8.22\% higher when using hallucinated descriptions generated by \texttt{GPT-4o}. Similar gains are observed with other generators: hallucinated descriptions from \texttt{Llama-3} and \texttt{ChemLLM} improve the average ROC-AUC for \texttt{Falcon3-Mamba} by 4.88\% and 6.01\%, respectively. These results demonstrate that speculative or loosely grounded hallucinations can, in some cases, be more effective than factual metadata.

Hallucinated descriptions often lead to substantial improvements over the \texttt{SMILES} baseline. Among the 21 model–generator combinations (7 models × 3 hallucination sources), 12 achieve higher average ROC-AUC than when using \texttt{SMILES}, with especially consistent gains observed for \texttt{Llama-3.1-8B} and \texttt{Falcon3-Mamba}, each of which benefits from all three hallucination sources. Even in cases where hallucinated descriptions do not outperform \texttt{SMILES}, they often remain competitive, suggesting their potential to convey useful inductive biases.

Similar patterns hold when compared to \texttt{MolT5}: 14 of 21 combinations achieve improved performance. In particular, hallucinations generated for \texttt{Llama-3.1-8B}, \texttt{Falcon3-Mamba}, and \texttt{GPT-4o} consistently surpass \texttt{MolT5}, highlighting that free-text generated by LLMs can rival or exceed specialized neural baselines. Overall, while not universally superior, hallucinated inputs frequently enhance performance, demonstrating their practical value for molecule property prediction.

Among all combinations, \texttt{Llama-3.1-8B} shows the largest gain: using hallucinations generated by \texttt{Llama-3-8B}, it achieves 15.80\% higher performance than \texttt{SMILES}, and 11.23\% over \texttt{MolT5}. \texttt{Falcon3-Mamba-7B} also surpasses all baselines by at least 4.88\%.

\textbf{These results suggest that hallucinated descriptions, despite being factually inconsistent, can enhance molecular classification performance in LLMs, sometimes even outperforming authoritative references.}

\subsection{Which LLMs Generate the Good Hallucinations?}

\begin{figure}[ht]
    \centering
    \includegraphics[width=1\linewidth]{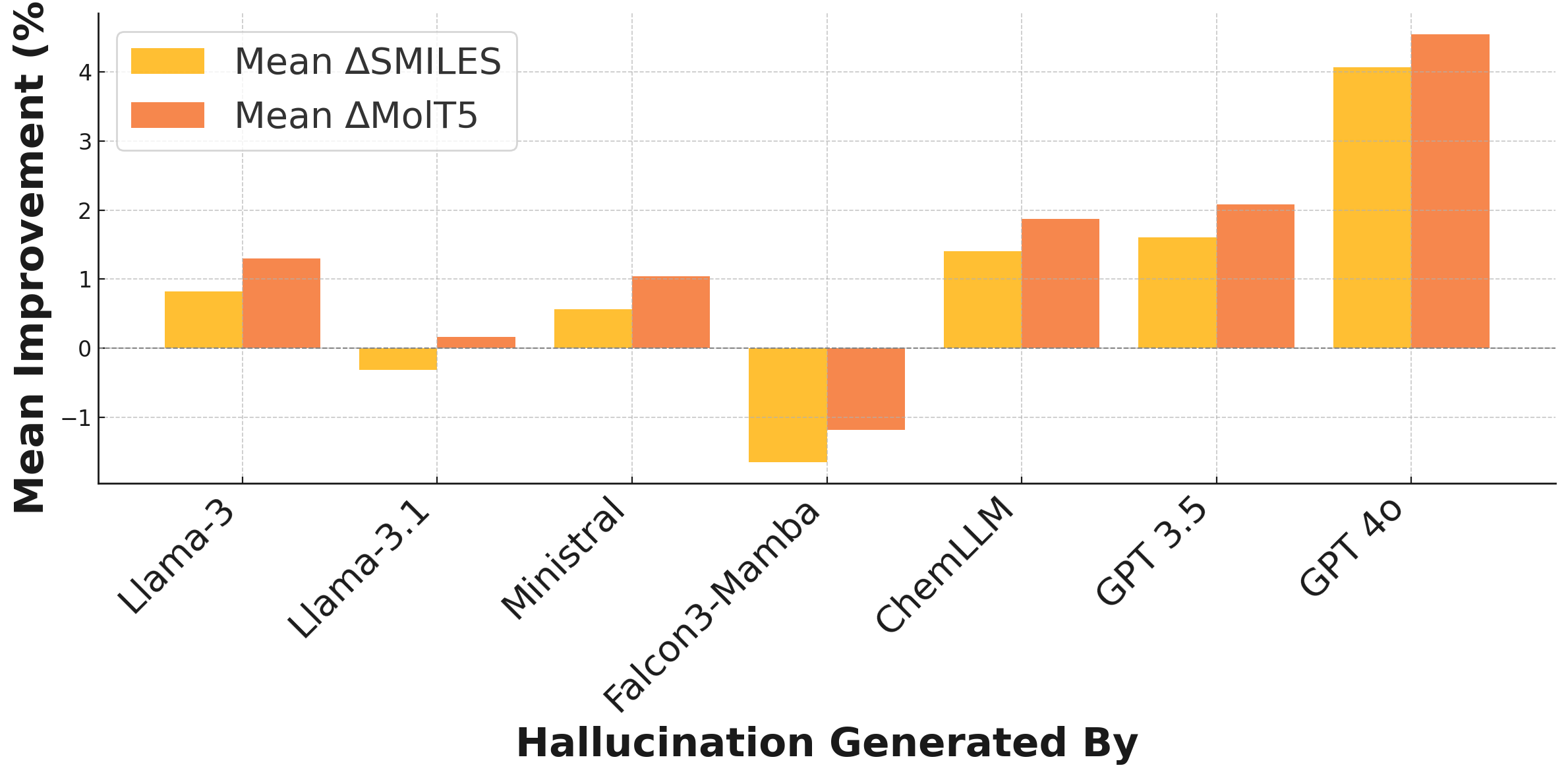}
    \caption{Average improvement in ROC-AUC across seven LLMs using hallucinations generated by different models. The x-axis indicates the source model of the hallucinations.}
    \label{figure:overall_hallu_improvement}
\end{figure}

To determine which LLMs produce the most useful hallucinations, we evaluate performance improvements across all seven models using hallucinations generated by different sources. Figure~\ref{figure:overall_hallu_improvement} presents the average gain in ROC-AUC when switching from \texttt{SMILES} or \texttt{MolT5} input to hallucinated descriptions.

Hallucinations from OpenAI models consistently yield the greatest improvements. Descriptions from \texttt{GPT-4o} improve average performance by 4.07\% over SMILES and 4.54\% over \texttt{MolT5}. \texttt{GPT-3.5} also provides solid gains of 1.6\% and 2.08\%, respectively. Among open-source models, hallucinations from \texttt{Llama-3} and \texttt{Mistral} offer modest improvements, while those from \texttt{Falcon3-Mamba} and \texttt{Llama-3.1} lead to slight decreases in performance.

\textbf{Overall, hallucinations generated by \texttt{GPT-4o} are the most beneficial across all models, suggesting that not all hallucinations are equally useful; type, quality, and coherence matter.}

\subsection{What Types of Hallucinations Are Most Useful?}

\begin{table}[htb!]
\centering
\scalebox{0.75}{
\begin{tabular}{lcccccc}
\toprule
\textbf{Source} & HIV & BBBP & Clintox & SIDER & Tox21 & \textbf{Avg.} \\
\midrule
\texttt{Llama-3} & 83.5 & 17.6 & 81.8 & 44.8 & 76.8 & 60.9 \\
\texttt{Llama-3.1} & 86.9 & 16.1 & 70.3 & 35.0 & 72.2 & 56.1 \\
\texttt{Mistral} & 87.6 & 16.1 & 60.1 & 37.1 & 74.1 & 55.0 \\
\texttt{Falcon3-Mamba} & 88.3 & 17.6 & 64.2 & 39.2 & 79.4 & 57.7 \\
\texttt{ChemLLM} & 44.2 & 27.3 & 33.8 & 35.0 & 19.9 & 32.0 \\
\texttt{GPT-3.5} & 92.2 & 15.6 & 37.8 & 24.5 & 77.4 & 49.5 \\
\texttt{GPT-4o} & 78.4 & 25.9 & 62.8 & 36.4 & 66.9 & 54.1 \\
\bottomrule
\end{tabular}}
\caption{Proportion (\%) of hallucinations that improve performance over all baselines for \texttt{Falcon3-Mamba-7B}, broken down by dataset and hallucination source.}
\label{tab:beneficial_hallucinations}
\end{table}

To understand the mechanism behind this improvement, we analyze a subset of \textit{beneficial hallucinations}, which are the ones that lead to correct predictions with greater confidence than any baseline. Focusing on \texttt{Falcon3-Mamba-7B}, we collect 18,872 such cases across five datasets and seven hallucination sources (Table~\ref{tab:beneficial_hallucinations}).

\begin{figure}[ht]
    \centering
    \includegraphics[width=1\linewidth]{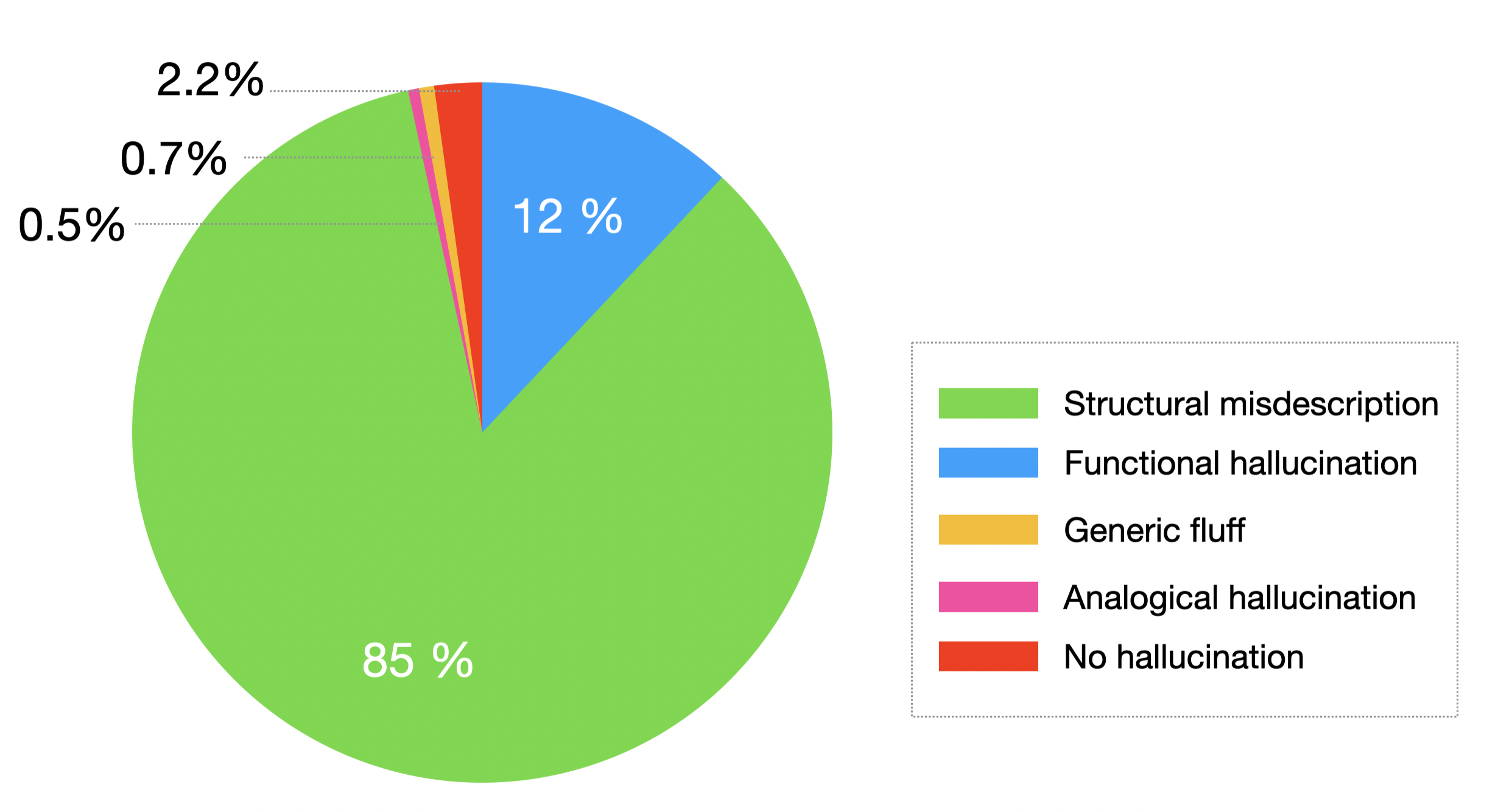}
    \caption{Distribution of hallucination types contributing to improved predictions for \texttt{Falcon3-Mamba-7B}.}
    \label{figure:pie_chart}
\end{figure}

Using the hallucination taxonomy established earlier, we find that the majority (85\%) of effective hallucinations are \textit{structural misdescriptions}, which are incorrect yet chemically plausible statements about the molecular structure. Another 12\% are \textit{functional hallucinations} involving inferred or speculative bioactivity. Generic or analogical hallucinations are rare and contribute little to performance (Figure~\ref{figure:pie_chart}).

\textbf{These hallucinations may act as \textit{implicit counterfactuals}}, offering alternative, albeit incorrect, chemical scenarios that prompt the LLM to refine its internal representation of the molecule. Rather than misleading the model, such perturbations may function as soft interventions, related to counterfactual reasoning in structured explainability frameworks. They often describe structural or functional properties that deviate from known molecular facts but nonetheless align with decision-relevant patterns.

This perspective suggests that hallucinations are not just noise: they reshape the model’s belief state by inducing hypothetical molecular interpretations, thereby enhancing its predictive confidence.

\FloatBarrier

\section{Ablation Studies}

To better understand when and why hallucinated text improves molecular prediction, we conduct ablation studies focusing on two factors: model size and generation temperature. We use the \texttt{Llama-3} family due to its consistent architecture and multiple available sizes. As \texttt{PubChem} coverage is incomplete, we report improvements relative to \texttt{SMILES} and \texttt{MolT5} baselines.

\subsection{Effect of Model Size}

Model size is a critical determinant of LLM performance~\citep{dubey2024llama}. We compare instruction-tuned versions of \texttt{Llama-3.2} (1B, 3B) and \texttt{Llama-3.1} (8B, 70B), using hallucinations generated by all other LLMs used for the experiments. For each model, we compute the average ROC-AUC improvement over the baselines.

\begin{figure}[ht]
    \centering
    \includegraphics[width=1\linewidth]{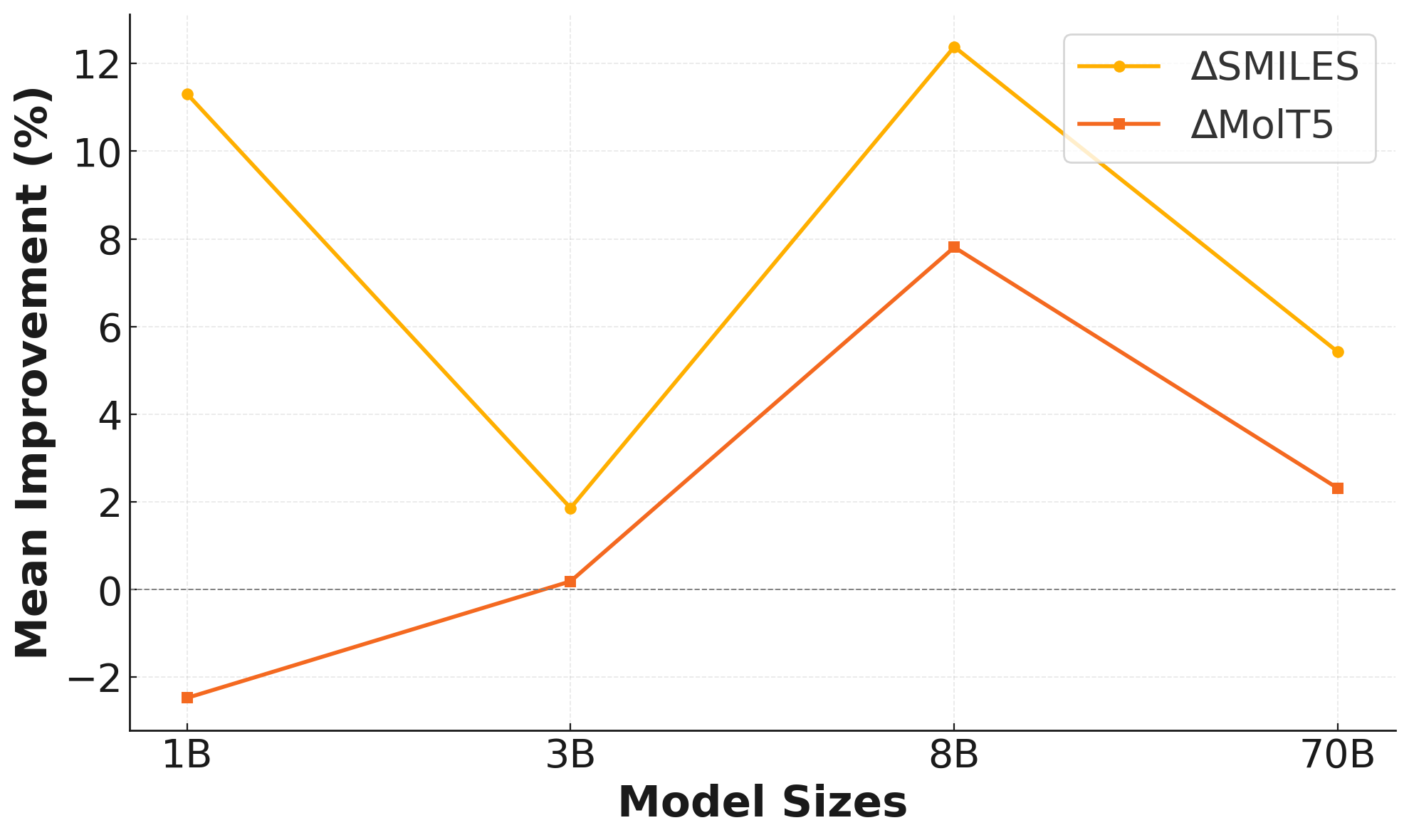}
    \caption{Average ROC-AUC improvement for \texttt{Llama-3} models of different sizes, using hallucinated descriptions.}
    \label{figure:model_size}
\end{figure}

As shown in Figure~\ref{figure:model_size}, performance improves steadily from 1B to 8B when compared to the \texttt{MolT5} baseline, suggesting that larger models better leverage the information, even when it is hallucinatory. The 70B model does not outperform the 8B model but still exceeds the smaller sizes. When compared to the \texttt{SMILES} baseline, all sizes show improvement.

\textbf{These findings suggest that larger models are more capable of extracting useful signals from hallucinated descriptions, though gains may plateau beyond 8B.}

\subsection{Effect of Generation Temperature}

\begin{figure}[ht]
    \centering
    \includegraphics[width=1\linewidth]{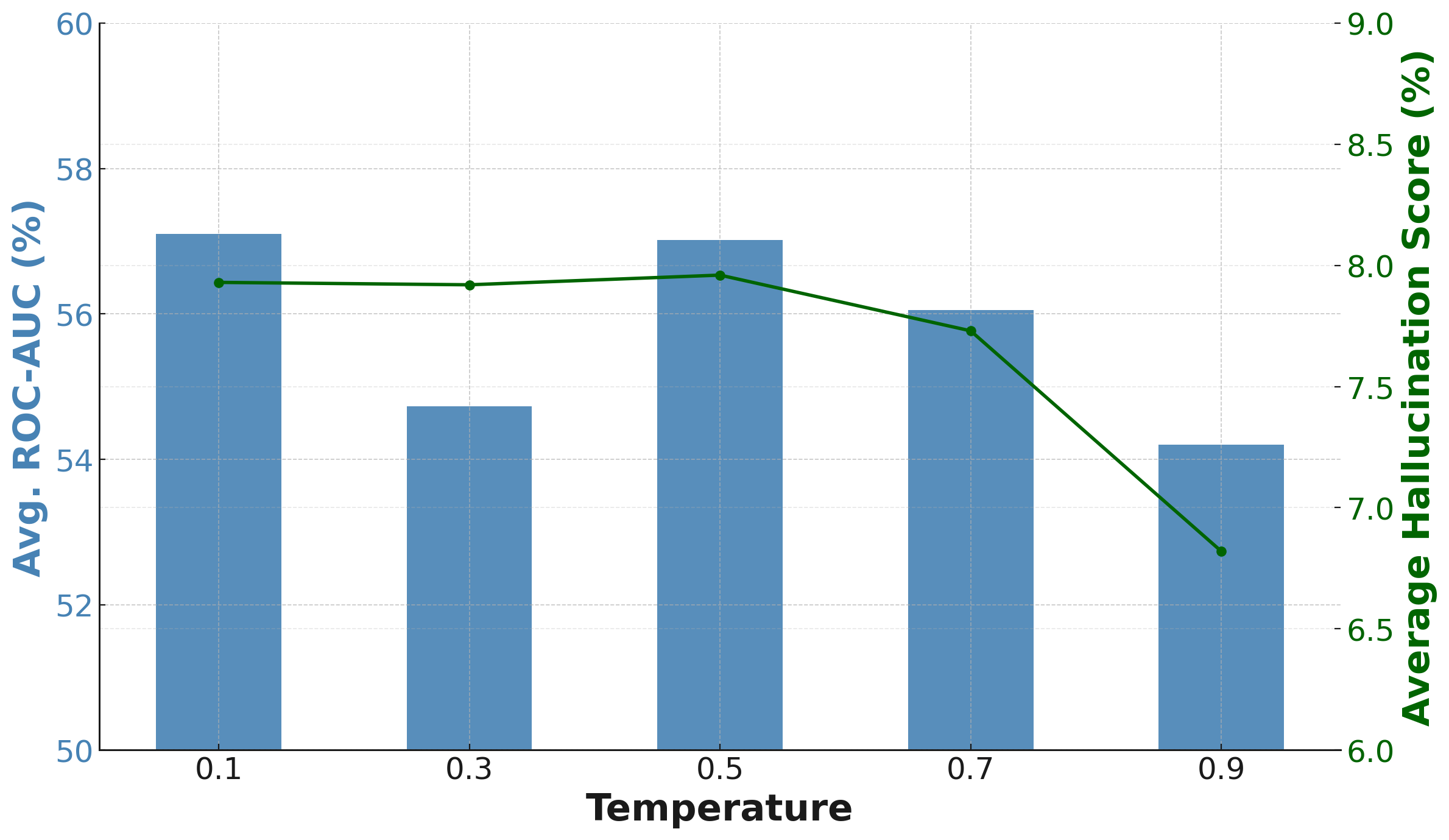}
    \caption{Average model performance and hallucination scores at different temperature settings (0.1–0.9) using \texttt{Llama-3.1-8B}.}
    \label{figure:temperature}
\end{figure}

Sampling temperature controls the randomness of generation~\citep{van2024random}, with higher temperatures producing more diverse and often less factual outputs~\citep{peeperkorn2024temperature,renze-2024-effect}. We vary the temperature between 0.1 and 0.9 while using \texttt{Llama-3.1-8B} to generate molecular descriptions, then assess both hallucination scores and downstream performance.

Figure~\ref{figure:temperature} shows that as the temperature increases, hallucination scores increase, that is, the factual consistency decreases. This aligns with prior expectations. However, the correlation with performance is non-monotonic: model accuracy peaks at temperatures 0.1 and 0.5, but dips slightly at 0.3 and 0.9. The difference across all temperatures is small (within ~3\%).

Notably, all temperature settings outperform both the \texttt{SMILES} (41.71\%) and \texttt{MolT5} (46.28\%) baselines (Table~\ref{tab:main_results}). \textbf{This confirms that even high-temperature, hallucinated text, despite reduced factual consistency, can still contribute positively to LLM performance.}

\section{Conclusion}

This paper investigates whether hallucinations can improve the performance of Large Language Models  (LLMs) in predicting molecule properties. We evaluate seven LLMs across five datasets by incorporating natural language descriptions of molecules, which often contain hallucinations, into the input prompts. 

Our results show that hallucinated input can improve classification performance for most LLMs, even outperforming accurate baselines like \texttt{MolT5} and \texttt{PubChem} in some cases. In particular, \texttt{Falcon3-Mamba-7B} and hallucinations generated by \texttt{GPT-4o} yield the most consistent improvements across datasets. 

We further categorize the beneficial hallucinations and find that \textit{structural misdescription} is the most common and impactful type, suggesting that factual deviation does not necessarily harm, and may even help, LLM decision-making. Ablation studies reveal that larger models are better at leveraging hallucinated descriptions, while variation in generation temperature has limited impact on downstream performance. 

Our findings offer a new perspective on the role of hallucinations in LLM-based scientific tasks, highlighting their potential to act as implicit counterfactuals that encourage flexible reasoning and robust prediction. Future work may further investigate the mechanisms by which LLMs benefit from such hallucinated input and how to leverage this creative potential in a safe and intentional manner, especially in high-stakes domains.

\newpage
\bibliography{aaai2026,anthology}


\clearpage



\appendix
\section{Details of the Models}
\label{sec:appendix_models}

In Table \ref{table:model_details}, we list the full names and the links of the LLMs used in the work. All the open-source LLMs can be applied directly using Transformers library by Huggingface.

\begin{table*}[ht]
\centering
\scalebox{0.7}{
\begin{tabular}{@{}lll@{}}
\toprule
\textbf{Model} & \textbf{Full-name} & \textbf{Link} \\ \midrule
\texttt{Llama-3-8B} & meta-llama/Meta-Llama-3-8B-Instruct & \url{https://huggingface.co/meta-llama/Llama-3.1-8B-Instruct} \\
\texttt{Llama-3.1-8B} & meta-llama/Llama-3.1-8B-Instruct & \url{https://huggingface.co/meta-llama/Llama-3.1-8B-Instruct} \\
\texttt{Ministral-8B} & mistralai/Ministral-8B-Instruct-2410 & \url{https://huggingface.co/mistralai/Ministral-8B-Instruct-2410} \\
\texttt{Falcon3-Mamba-7B} & tiiuae/Falcon3-Mamba-7B-Instruct & \url{https://huggingface.co/tiiuae/Falcon3-Mamba-7B-Instruct} \\
\texttt{ChemLLM-7B} & AI4Chem/ChemLLM-7B-Chat-1\_5-DPO & \url{https://huggingface.co/AI4Chem/ChemLLM-7B-Chat-1_5-DPO} \\
\texttt{GPT-3.5} & gpt-3.5-turbo & - \\
\texttt{GPT-4o} & gpt-4o-2024-08-06 & - \\ \bottomrule
\end{tabular}}
\caption{Details of the evaluated models, including their full names and links to their respective HuggingFace pages.}
\label{table:model_details}
\end{table*}

\section{Details of the Datasets}\label{sec:appendix_dataset}

The details, including the number of samples and the count of positive labels in each dataset are reported in Table \ref{table:datasets_detail}.

\begin{table}[ht]
\centering
\begin{tabular}{@{}lrr@{}}
\toprule
\textbf{Dataset} & \textbf{Number} & \textbf{Positive Label} \\ \midrule
HIV     & 4113 & 175 \\
BBBP    & 205  & 172 \\
Clintox & 148  & 11  \\
SIDER   & 143  & 74  \\
Tox21   & 783  & 108 \\ \bottomrule
\end{tabular}
\caption{Summary of the datasets used, including the number of samples and the count of positive labels in each dataset.}
\label{table:datasets_detail}
\end{table}

\section{Prompt Template for Different Tasks}\label{sec:appendix_prompt_template}

For each task, we use different instruction as they aim for predicting different properties of molecules.

\subsection{HIV}
\begin{tcolorbox}[colback=skyblue!20, colframe=skyblue!80!black, width=0.48\textwidth, rounded corners]
\textit{System}: You are an expert in drug discovery.\\
\textit{User}: $[SMILES]$\;$[Description]$\\
Does the molecule have the ability to inhibit HIV replication? Only answer Yes or No:
\end{tcolorbox}

\subsection{BBBP}
\begin{tcolorbox}[colback=skyblue!20, colframe=skyblue!80!black, width=0.48\textwidth, rounded corners]
\textit{System}: You are an expert in drug discovery.\\
\textit{User}: $[SMILES]$\;$[Description]$\\
Does the molecule have the ability to penetrate the blood-brain barrier? Only answer Yes or No:
\end{tcolorbox}

\subsection{Clintox}
\begin{tcolorbox}[colback=skyblue!20, colframe=skyblue!80!black, width=0.48\textwidth, rounded corners]
\textit{System}: You are an expert in drug discovery.\\
\textit{User}: $[SMILES]$\;$[Description]$\\
Did the molecule fail clinical trials due to toxicity? Only answer Yes or No:
\end{tcolorbox}

\subsection{SIDER}
\begin{tcolorbox}[colback=skyblue!20, colframe=skyblue!80!black, width=0.48\textwidth, rounded corners]
\textit{System}: You are an expert in drug discovery.\\
\textit{User}: $[SMILES]$\;$[Description]$\\
Does the molecule cause side effects on the reproductive system or breast? Only answer Yes or No:
\end{tcolorbox}

\subsection{Tox21}
\begin{tcolorbox}[colback=skyblue!20, colframe=skyblue!80!black, width=0.48\textwidth, rounded corners]
\textit{System}: You are an expert in drug discovery.\\
\textit{User}: $[SMILES]$\;$[Description]$\\
Does the molecule have the potential toxicity affecting mitochondrial membrane potential (SR-MMP)? Only answer Yes or No:
\end{tcolorbox}

\section{Hallucination Score for LLMs Generated Description}\label{section:appendix_hallu_score}

We evaluate the LLMs generated description using HHM-2.1-Open and report the results in Table \ref{tab:hallucination_score_overal}.

\begin{table*}[htb]
    \centering
    \begin{tabular}{lcccccc}
         \textbf{Model}&  \textbf{HIV}&  \textbf{BBBP}&  \textbf{Clintox} &  \textbf{SIDER}&  \textbf{Tox21}& \textbf{Avg}\\
         \toprule
         \texttt{Llama-3.1-8B}&  7.12&  7.76&  8.67&  9.45&  6.86& 7.97\\
         \texttt{Llama-3-8B} &  7.15&  8.04&  7.24&  7.57&  7.11& 7.42\\
         \texttt{Ministral-8B} &  14.45&  13.47&  14.23&  13.17&  12.56& 13.58\\
         \texttt{Falcon3-Mamba-7B}&  8.74&  9.05&  8.92&  11.13&  9.26& 9.42\\
         \texttt{ChemLLM-7B} &  20.48&  17.75&  23.36&  22.20&  20.66& 20.89\\
         \texttt{GPT-3.5} &  8.62&  8.35&  6.85&  7.20&  6.90& 7.58\\
         \texttt{GPT-4o} &  8.51&  7.67&  8.51&  7.28&  7.44& 7.88\\
    \bottomrule
    \end{tabular}
    \caption{The HHM-2.1 score \% for LLMs across five datasets, all the models are using the same hyperparameters, e.g.: the temperature, max new tokens.}
    \label{tab:hallucination_score_overal}
\end{table*}

\section{Full Results of Main Experiments}\label{sec:appendix_full_reults}

The full results for all LLMs used in our study are presented in Table \ref{table:full_results_all_LLMs}. Each LLM is evaluated using at least nine prompt templates: two from the baselines and seven incorporating hallucinations generated by other models.

\begin{table*}[ht]
\centering
\scalebox{0.8}{
\begin{tabular}{llccccccccc}
\toprule
\textbf{Model}                          & $\boldsymbol{[Description]}$ & \textbf{HIV}       & \textbf{BBBP}    & \textbf{Clintox} & \textbf{SIDER}   & \textbf{Tox21}   & \textbf{Avg}     & $\boldsymbol\Delta$\textbf{\texttt{SMILES}} & $\boldsymbol\Delta$\textbf{\texttt{MolT5}} & $\boldsymbol\Delta$\textbf{\texttt{PubChem}}   \\ 
\midrule
\multirow{10}{*}{\textbf{\texttt{Llama-3-8B}}}      & \texttt{SMILES}      & \textbf{67.78} & 53.08          & \underline{63.04} & \textbf{61.79} & 60.34          & 61.21   & -       & 6.53    & 0.55    \\
                                & \texttt{MolT5}       & 47.65          & 59.65          & 43.20          & \underline{61.14} & 61.73          & 54.68   & -6.53   & -       & -5.98   \\
                                & \texttt{PubChem}     & 49.85          & \textbf{72.48} & 53.35          & 59.99          & \textbf{67.60} & 60.65   & -0.55   & 5.98    & -       \\
                                & \texttt{Llama-3}     & 53.13          & 58.21          & 57.13          & 57.07          & 59.75          & 57.06   & -4.15   & 2.38    & -3.59   \\
                                & \texttt{Llama-3.1}   & 57.60          & 51.02          & 39.75          & 60.63          & 59.70          & 53.74   & -7.46   & -0.93   & -6.91   \\
                                & \texttt{Ministral}   & 55.09          & \underline{61.13} & \textbf{63.84} & 57.56          & 55.41          & 58.61   & -2.60   & 3.93    & -2.04   \\
                                & \texttt{Falcon3}     & 54.02          & 51.73          & 43.00          & 47.06          & 48.91          & 48.94   & -12.26  & -5.73   & -11.71  \\
                                & \texttt{ChemLLM}     & 46.83          & 60.22          & 52.36          & 58.50          & 50.83          & 53.75   & -7.46   & -0.93   & -6.90   \\
                                & \texttt{GPT-3.5}     & \underline{58.45} & 46.52          & 57.13          & 56.15          & \underline{66.00} & 56.85   & -4.36   & 2.18    & -3.80   \\
                                & \texttt{GPT-4o}      & 45.97          & 53.72          & 54.81          & 45.86          & 60.90          & 52.25   & -8.96   & -2.42   & -8.40   \\
\midrule
\multirow{10}{*}{\textbf{\texttt{Llama-3.1-8B}}}    & \texttt{SMILES}      & 38.10          & 37.56          & 35.30          & 52.70          & 44.89          & 41.71   & -       & -4.57   & -22.36  \\
                                & \texttt{MolT5}       & 43.04          & 45.30          & 39.28          & 52.06          & 51.72          & 46.28   & 4.57    & -       & -17.8   \\
                                & \texttt{PubChem}     & 58.94          & \textbf{71.42} & 66.03          & \textbf{57.54} & \textbf{66.45} & 64.08   & 22.36   & 17.80   & -       \\
                                & \texttt{Llama-3}     & 56.80          & 47.71          & 68.08          & 54.21          & 60.75          & 57.51   & 15.80   & 11.23   & -6.57   \\
                                & \texttt{Llama-3.1}   & 50.92          & 42.11          & 49.97          & 38.11          & 51.74          & 46.57   & 4.86    & 0.29    & -17.51  \\
                                & \texttt{Ministral}   & \underline{59.73} & 50.18          & \underline{70.01} & 51.65          & 55.59          & 57.43   & 15.72   & 11.15   & -6.65   \\
                                & \texttt{Falcon3}     & 57.92          & 51.11          & 53.22          & 42.38          & 47.57          & 50.44   & 8.73    & 4.16    & -13.64  \\
                                & \texttt{ChemLLM}     & 50.83          & 52.18          & 47.45          & 53.17          & 49.84          & 50.69   & 8.98    & 4.41    & -13.39  \\
                                & \texttt{GPT-3.5}     & \textbf{64.66} & 47.20          & \textbf{73.19} & \underline{55.57} & 59.70          & 60.07   & 18.35   & 13.79   & -4.01   \\
                                & \texttt{GPT-4o}      & 56.64          & \underline{53.37} & 55.28          & 52.60          & \underline{61.65} & 55.91   & 14.20   & 9.63    & -8.17   \\
\midrule
\multirow{10}{*}{\textbf{\texttt{Ministral-8B}}}    & \texttt{SMILES}      & \underline{59.35} & 48.87          & 60.19          & 57.27          & 57.42          & 56.62   & -       & 3.53    & -5.63   \\
                                & \texttt{MolT5}       & 44.54          & 44.10          & 53.95          & \underline{63.83} & 59.02          & 53.09   & -3.53   & -       & -9.16   \\
                                & \texttt{PubChem}     & 50.08          & \underline{52.77} & \textbf{79.23} & \textbf{64.45} & \textbf{64.72} & 62.25   & 5.63    & 9.16    & -       \\
                                & \texttt{Llama-3}     & 52.91          & 31.27          & \underline{65.49} & 50.45          & 55.81          & 51.19   & -5.43   & -1.90   & -11.06  \\
                                & \texttt{Llama-3.1}   & 56.06          & 46.62          & 36.23          & 55.95          & 58.33          & 50.64   & -5.98   & -2.45   & -11.61  \\
                                & \texttt{Ministral}   & \textbf{60.04} & 46.42          & 55.34          & 52.82          & 53.58          & 53.64   & -2.98   & 0.56    & -8.61   \\
                                & \texttt{Falcon3}     & 55.37          & 50.16          & 53.62          & 47.14          & 58.30          & 52.92   & -3.70   & -0.17   & -9.33   \\
                                & \texttt{ChemLLM}     & 46.35          & 48.66          & 62.11          & 58.97          & 55.36          & 54.29   & -2.33   & 1.20    & -7.96   \\
                                & \texttt{GPT-3.5}     & 53.23          & 36.89          & 46.85          & 51.63          & 59.86          & 49.69   & -6.93   & -3.40   & -12.56  \\
                                & \texttt{GPT-4o}      & 51.66          & \textbf{64.94} & 58.73          & 60.54          & \underline{61.07} & 59.39   & 2.77    & 6.30    & -2.86   \\
\midrule
\multirow{10}{*}{\textbf{\texttt{Falcon3-Mamba-7B}}}& \texttt{SMILES}      & 40.64          & 48.33          & 31.72          & 52.53          & 47.45          & 44.13   & -       & 0.21    & -1.34   \\
                                & \texttt{MolT5}       & 49.02          & 47.92          & 18.58          & 55.84          & 48.27          & 43.93   & -0.21   & -       & -1.54   \\
                                & \texttt{PubChem}     & 45.40          & 58.84          & 24.55          & 52.00          & 46.56          & 45.47   & 1.34    & 1.54    & -       \\
                                & \texttt{Llama-3}     & 47.84          & 53.40          & \underline{50.76} & 51.31          & 48.44          & 50.35   & 6.22    & 6.42    & 4.88    \\
                                & \texttt{Llama-3.1}   & 48.36          & \underline{60.15} & 34.84          & \textbf{57.60} & \textbf{54.26} & 51.04   & 6.91    & 7.12    & 5.57    \\
                                & \texttt{Ministral}   & 47.32          & 47.82          & 27.94          & 51.90          & 47.54          & 44.50   & 0.37    & 0.58    & -0.97   \\
                                & \texttt{Falcon3}     & 44.59          & 57.05          & 39.42          & 50.10          & 46.64          & 47.56   & 3.43    & 3.63    & 2.09    \\
                                & \texttt{ChemLLM}     & \textbf{53.46} & \textbf{61.82} & 45.12          & 53.56          & 43.43          & 51.48   & 7.35    & 7.55    & 6.01    \\
                                & \texttt{GPT-3.5}     & \underline{51.63} & 48.47          & 45.26          & 50.98          & 46.73          & 48.61   & 4.48    & 4.69    & 3.14    \\
                                & \texttt{GPT-4o}      & 51.59          & 55.73          & \textbf{53.55} & \underline{56.03} & \underline{51.54} & 53.69   & 9.55    & 9.76    & 8.22    \\
\midrule
\multirow{10}{*}{\textbf{\texttt{ChemLLM-7B}}}      & \texttt{SMILES}      & 55.54          & 38.69          & 24.02          & 47.85          & \textbf{65.59} & 46.34   & -       & -1.93   & -8.36   \\
                                & \texttt{MolT5}       & 50.33          & 41.01          & 35.43          & \underline{53.06} & 61.50          & 48.27   & 1.93    & -       & -6.43   \\
                                & \texttt{PubChem}     & 54.96          & \textbf{72.27} & 34.37          & 51.08          & 60.79          & 54.69   & 8.36    & 6.43    & -       \\
                                & \texttt{Llama-3}     & 56.29          & 45.68          & \underline{47.51} & 44.95          & 51.59          & 49.20   & 2.87    & 0.94    & -5.49   \\
                                & \texttt{Llama-3.1}   & 57.17          & 43.02          & 28.33          & 43.83          & 56.96          & 45.86   & -0.47   & -2.40   & -8.83   \\
                                & \texttt{Ministral}   & \textbf{61.35} & 37.70          & 39.95          & 37.76          & 51.64          & 45.68   & -0.66   & -2.59   & -9.01   \\
                                & \texttt{Falcon3}     & \underline{57.52} & 39.76          & 29.46          & 44.71          & 50.18          & 44.33   & -2.01   & -3.94   & -10.36  \\
                                & \texttt{ChemLLM}     & 45.37          & 41.83          & 42.80          & \textbf{53.15} & 46.60          & 45.95   & -0.39   & -2.32   & -8.74   \\
                                & \texttt{GPT-3.5}     & 57.48          & 38.16          & \textbf{51.76} & 41.72          & 59.37          & 49.70   & 3.36    & 1.43    & -4.99   \\
                                & \texttt{GPT-4o}      & 52.68          & \underline{55.73} & 40.74          & 44.59          & \underline{62.63} & 51.28   & 4.94    & 3.01    & -3.42   \\
\midrule
\multirow{10}{*}{\textbf{\texttt{GPT-3.5}}}        & \texttt{SMILES}      & \textbf{61.32} & 28.94          & 63.90          & 53.77          & 61.92          & 53.97   & -       & -2.56   & -11.01  \\
                                & \texttt{MolT5}       & 49.34          & \underline{52.64} & \underline{64.30} & 56.00          & 60.39          & 56.53   & 2.56    & -       & -8.44   \\
                                & \texttt{PubChem}     & 56.64          & \textbf{61.11} & \textbf{76.34} & 59.99          & \textbf{70.81} & 64.98   & 11.01   & 8.44    & -       \\
                                & \texttt{Llama-3}     & 52.55          & 30.00          & 38.27          & 52.22          & 55.71          & 45.75   & -8.22   & -10.78  & -19.23  \\
                                & \texttt{Llama-3.1}   & 55.90          & 39.31          & 41.83          & 57.16          & 59.60          & 50.76   & -3.21   & -5.77   & -14.22  \\
                                & \texttt{Ministral}   & 55.84          & 29.63          & 39.89          & 56.61          & 56.31          & 47.66   & -6.31   & -8.88   & -17.32  \\
                                & \texttt{Falcon3}     & 52.11          & 27.96          & 54.29          & 48.69          & 59.10          & 48.43   & -5.54   & -8.10   & -16.55  \\
                                & \texttt{ChemLLM}     & 50.27          & 45.97          & 63.03          & 53.77          & 54.57          & 53.52   & -0.45   & -3.01   & -11.46  \\
                                & \texttt{GPT-3.5}     & \underline{58.47} & 33.53          & 44.81          & \underline{62.24} & 56.46          & 51.10   & -2.87   & -5.43   & -13.88  \\
                                & \texttt{GPT-4o}      & 49.67          & 34.61          & 60.83          & \textbf{62.41} & \underline{65.36} & 54.57   & 0.60    & -1.96   & -10.40  \\
\midrule
\multirow{10}{*}{\textbf{\texttt{GPT-4o}}}         & \texttt{SMILES}      & 47.19          & 46.79          & 40.00          & 52.30          & 63.25          & 49.91   & -       & 2.10    & -6.17   \\
                                & \texttt{MolT5}       & 41.73          & 40.57          & 38.09          & \textbf{65.22} & 53.42          & 47.80   & -2.10   & -       & -8.27   \\
                                & \texttt{PubChem}     & 50.96          & 50.34          & \underline{56.67} & \underline{56.89} & \textbf{65.52} & 56.08   & 6.17    & 8.27    & -       \\
                                & \texttt{Llama-3}     & 48.95          & 36.23          & 52.97          & 42.73          & 62.00          & 48.58   & -1.33   & 0.77    & -7.5    \\
                                & \texttt{Llama-3.1}   & 53.02          & 45.57          & 54.10          & 48.71          & 64.07          & 53.09   & 3.19    & 5.29    & -2.99   \\
                                & \texttt{Ministral}   & \textbf{55.34} & 40.13          & 49.46          & 42.44          & 64.17          & 50.31   & 0.40    & 2.50    & -5.77   \\
                                & \texttt{Falcon3}     & \underline{54.01} & 35.37          & 50.00          & 45.83          & 63.34          & 49.71   & -0.20   & 1.90    & -6.37   \\
                                & \texttt{ChemLLM}     & 53.00          & \textbf{54.92} & 55.59          & 45.36          & 61.15          & 54.00   & 4.10    & 6.20    & -2.08   \\
                                & \texttt{GPT-3.5}     & 47.51          & 40.80          & 46.03          & 47.27          & 63.82          & 49.09   & -0.82   & 1.28    & -6.99   \\
                                & \texttt{GPT-4o}      & 53.30          & \underline{52.73} & \textbf{57.12} & 47.82          & \underline{65.34} & 55.26   & 5.35    & 7.46    & -0.82   \\
\bottomrule
\end{tabular}}
\caption{Full results (ROC-AUC \%) for all LLMs, including hallucinations generated by each model and baselines. $\boldsymbol{[Description]}$ indicates the source of the hallucination or baseline used.}\label{table:full_results_all_LLMs}
\label{tab:main_results}
\end{table*}

\begin{table*}[ht]
\centering
\scalebox{0.75}{
\begin{tabular}{llccccccccc}
\toprule
\textbf{Model} & $\boldsymbol{[Description]}$ & \textbf{HIV} & \textbf{BBBP} & \textbf{Clintox} & \textbf{SIDER} & \textbf{Tox21} & \textbf{Avg} & $\boldsymbol\Delta$\textbf{\texttt{SMILES}} & $\boldsymbol\Delta$\textbf{\texttt{MolT5}} & $\boldsymbol\Delta$\textbf{\texttt{PubChem}} \\
\midrule
\multirow{6}{*}{\textbf{\texttt{Llama-3-8B}}} & \texttt{SMILES}    & \textbf{67.78} & 53.08          & \textbf{63.04} & \textbf{61.79} & 60.34          & \textbf{61.21} & \textbf{-}      & \textbf{6.53}   & \textbf{0.56}    \\
                                & \texttt{MolT5}     & 47.65          & 59.65          & 43.20          & \underline{61.14} & 61.73          & 54.68          & -6.53           & -               & -5.97            \\
                                & \texttt{PubChem}   & 49.85          & \textbf{72.48} & 53.35          & 59.99          & \textbf{67.60} & \underline{60.65} & \underline{-0.56} & \underline{5.97}   & \underline{-}       \\
                                & \texttt{Llama Family} & 57.60          & 58.21          & 57.13          & 60.63          & 59.75          & 58.66          & -2.55           & 3.98            & -1.99            \\
                                & \texttt{GPT Family}   & \underline{58.45} & 53.72          & \underline{57.13} & 56.15          & \underline{66.00} & 58.29          & -2.92           & 3.61            & -2.36            \\
                                & \texttt{ChemLLM}   & 46.83          & \underline{60.22} & 52.36          & 58.50          & 50.83          & 53.75          & -7.46           & -0.93           & -6.90            \\
\midrule
\multirow{6}{*}{\textbf{\texttt{Llama-3.1-8B}}} & \texttt{SMILES}    & 38.10          & 37.56          & 35.30          & 52.70          & 44.89          & 41.71          & -               & -4.57           & -22.37           \\
                                & \texttt{MolT5}     & 43.04          & 45.30          & 39.28          & 52.06          & 51.72          & 46.28          & 4.57            & -               & -17.80           \\
                                & \texttt{PubChem}   & 58.94          & \textbf{71.42} & 66.03          & \textbf{57.54} & \textbf{66.45} & \textbf{64.08} & \textbf{22.37}  & \textbf{17.80}  & \textbf{-}       \\
                                & \texttt{Llama Family} & 56.80          & 47.71          & \textbf{68.08} & \underline{54.21} & 60.75          & 57.51          & 15.80           & 11.23           & -6.57            \\
                                & \texttt{GPT Family}   & \textbf{64.66} & \underline{53.37} & \underline{66.19} & 55.57          & \underline{61.65} & \underline{61.69} & \underline{19.98}  & \underline{15.41}  & \underline{-2.39}   \\
                                & \texttt{ChemLLM}   & 50.83          & 52.18          & 47.45          & 53.17          & 49.84          & 50.69          & 8.98            & 4.41            & -13.39           \\
\midrule
\multirow{6}{*}{\textbf{\texttt{Ministral-8B}}} & \texttt{SMILES}    & \underline{59.35} & 48.87          & 60.19          & 57.27          & 57.42          & 56.62          & -               & 3.53            & -5.63            \\
                                & \texttt{MolT5}     & 44.54          & 44.10          & 53.95          & \textbf{63.83} & 59.02          & 53.09          & -3.53           & -               & -9.16            \\
                                & \texttt{PubChem}   & 50.08          & 52.77          & \textbf{79.23} & \underline{63.15} & \textbf{64.72} & \textbf{62.25} & \textbf{5.63}   & \textbf{9.16}   & \textbf{-}       \\
                                & \texttt{Llama Family} & 56.06          & 46.62          & \underline{65.49} & 55.95          & 58.33          & 56.49          & -0.13           & 3.40            & -5.76            \\
                                & \texttt{GPT Family}   & 53.23          & \textbf{64.94} & 58.73          & 60.54          & \underline{61.07} & \underline{59.70} & \underline{3.08}   & \underline{6.61}   & \underline{-2.55}   \\
                                & \texttt{ChemLLM}   & 46.35          & \underline{48.66} & 62.11          & 58.97          & 55.36          & 54.29          & -2.33           & 1.20            & -7.96            \\
\midrule
\multirow{6}{*}{\textbf{\texttt{Falcon3-Mamba-7B}}} & \texttt{SMILES}    & 40.64          & 48.33          & 31.72          & 52.53          & 47.45          & 44.13          & -               & 0.20            & -1.34            \\
                                & \texttt{MolT5}     & 49.02          & 47.92          & 18.58          & \underline{55.84} & 48.27          & 43.93          & -0.20           & -               & -1.54            \\
                                & \texttt{PubChem}   & 45.40          & 58.84          & 24.55          & 52.00          & 46.56          & 45.47          & 1.34            & 1.54            & -                \\
                                & \texttt{Llama Family} & 50.70          & \underline{60.15} & \underline{50.76} & \textbf{57.60} & \textbf{54.26} & \textbf{54.23} & \textbf{10.10}  & \textbf{10.30}  & \textbf{8.76}    \\
                                & \texttt{GPT Family}   & 51.63          & 55.73          & \textbf{53.55} & 56.03          & \underline{51.54} & \underline{53.70} & \underline{9.57}   & \underline{9.77}   & \underline{8.23}    \\
                                & \texttt{ChemLLM}   & \textbf{53.46} & \textbf{61.82} & 45.12          & 53.56          & 43.43          & 51.48          & 7.35            & 7.55            & 6.01             \\
\midrule
\multirow{6}{*}{\textbf{\texttt{ChemLLM-7B}}} & \texttt{SMILES}    & 55.54          & 38.69          & 24.02          & 47.85          & \textbf{65.59} & 46.34          & -               & -1.93           & -8.35            \\
                                & \texttt{MolT5}     & 50.33          & 41.01          & 35.43          & \underline{53.06} & 61.50          & 48.27          & 1.93            & -               & -6.42            \\
                                & \texttt{PubChem}   & 54.96          & \textbf{72.27} & 34.37          & 51.08          & 60.79          & \textbf{54.69} & \textbf{8.35}   & \textbf{6.42}   & \textbf{-}       \\
                                & \texttt{Llama Family} & 57.17          & 45.68          & 47.51          & 44.95          & 56.96          & 50.45          & 4.11            & 2.18            & -4.24            \\
                                & \texttt{GPT Family}   & \textbf{57.48} & \underline{55.73} & \textbf{51.76} & 44.59          & \underline{62.63} & \underline{54.44} & \underline{8.10}   & \underline{6.17}   & \underline{-0.25}   \\
                                & \texttt{ChemLLM}   & 45.37          & 41.83          & \underline{42.80} & \textbf{53.15} & 46.60          & 45.95          & -0.39           & -2.32           & -8.74            \\
\midrule
\multirow{6}{*}{\textbf{\texttt{GPT-3.5}}} & \texttt{SMILES}    & \textbf{61.32} & 28.94          & 63.90          & 53.77          & 61.92          & 53.97          & -               & -2.56           & -11.01           \\
                                & \texttt{MolT5}     & 49.34          & \underline{52.64} & \underline{64.30} & 56.00          & 60.39          & \underline{56.53} & \underline{2.56}   & -               & -8.45            \\
                                & \texttt{PubChem}   & 56.64          & \textbf{61.11} & \textbf{76.34} & \underline{62.41} & \textbf{70.81} & \textbf{64.98} & \textbf{11.01}  & \textbf{8.44}   & \textbf{-}       \\
                                & \texttt{Llama Family} & 55.90          & 39.31          & 41.83          & 57.16          & 59.60          & 50.76          & -3.21           & -5.77           & -14.22           \\
                                & \texttt{GPT Family}   & \underline{58.47} & 34.61          & 60.83          & \textbf{62.41} & \underline{65.36} & 56.34          & 2.37            & -0.19           & -8.64            \\
                                & \texttt{ChemLLM}   & 50.27          & 45.97          & 63.03          & 53.77          & 54.57          & 53.52          & -0.45           & -3.01           & -11.46           \\
\midrule
\multirow{6}{*}{\textbf{\texttt{GPT-4o}}} & \texttt{SMILES}    & 47.19          & 46.79          & 40.00          & 52.30          & 63.25          & 49.91          & -               & 2.11            & -6.17            \\
                                & \texttt{MolT5}     & 41.73          & 40.57          & 38.09          & \textbf{65.22} & 53.42          & 47.80          & -2.11           & -               & -8.28            \\
                                & \texttt{PubChem}   & 50.96          & 50.34          & 56.67          & \underline{56.89} & \textbf{65.52} & \textbf{56.08} & \textbf{6.17}   & \textbf{8.28}   & \textbf{-}       \\
                                & \texttt{Llama Family} & 53.02          & 45.57          & 54.10          & 48.71          & 64.07          & 53.09          & 3.18            & 5.29            & -2.99            \\
                                & \texttt{GPT Family}   & \textbf{53.30} & 52.73          & \textbf{57.12} & 47.82          & \underline{65.34} & \underline{55.26} & \underline{5.35}   & \underline{7.46}   & \underline{-0.82}   \\
                                & \texttt{ChemLLM}   & \underline{53.00} & \textbf{54.92} & \underline{55.59} & 45.36          & 61.15          & 54.00          & 4.09            & 6.20            & -2.08            \\
\bottomrule
\end{tabular}}
\caption{Curated results with combined "Llama Family" and "GPT Family" descriptions. For these families, the best performance across their respective constituent models is used for each dataset. Within each model's group, the best and second-best scores per column are highlighted in \textbf{bold} and \underline{underlined}, respectively.}
\label{tab:final_curated_results}
\end{table*}

\section{Full Results for Model Size Experiments}\label{sec:appendix_model_size}

The full results for the model size experiments are shown in Table \ref{table:full_results_model_size}. Additionally, the hallucination scores for each model size are provided in Table \ref{tab:hallucination_model_size}.

\begin{table*}[ht]
\centering
\scalebox{0.7}{
\begin{tabular}{ccccccccc}
$\boldsymbol{[Description]}$ & \textbf{HIV} & \textbf{BBBP} & \textbf{Clintox} & \textbf{SIDER} & \textbf{Tox21} & \textbf{Avg} & $\boldsymbol\Delta$\textbf{ \texttt{SMILES}} & $\boldsymbol\Delta$\textbf{ \texttt{MolT5}}  \\ 
\toprule
\multicolumn{9}{c}{\textbf{\texttt{Llama-3.2-1B}}}                                                                                                                                                                          \\ 
\midrule
 \texttt{SMILES}                       & 44.27       & 32.31        & 19.31           & 60.79         & 46.88         & 40.71       & 0.00                              & -13.77                            \\
\texttt{MolT5}                         & 45.26       & 38.30        & 79.50           & 57.95         & 51.42         & 54.49       & 13.77                             & 0.00                              \\
\texttt{Llama-3}                     & 46.79       & 47.32        & 60.19           & 56.23         & 44.25         & 50.96       & 10.24                             & -3.53                             \\
\texttt{Llama-3.1}                   & 44.79       & 50.93        & 51.56           & 59.09         & 53.73         & 52.02       & 11.31                             & -2.47                             \\
\texttt{Ministral}                   & 46.94       & 42.00        & 54.48           & 51.94         & 52.71         & 49.61       & 8.90                              & -4.87                             \\
\texttt{Falcon3}                     & 52.08       & 58.58        & 56.01           & 50.55         & 47.14         & 52.87       & 12.16                             & -1.62                             \\
\texttt{ChemLLM}                      & 46.23       & 44.77        & 60.25           & 56.35         & 44.19         & 50.36       & 9.65                              & -4.13                             \\
\texttt{GPT-3.5}                      & 54.33       & 45.24        & 58.00           & 56.01         & 58.94         & 54.50       & 13.79                             & 0.02                              \\
\texttt{GPT-4o}                       & 50.29       & 60.02        & 64.43           & 52.88         & 52.10         & 55.94       & 15.23                             & 1.46                              \\ 
\midrule
\multicolumn{9}{c}{\textbf{\texttt{Llama-3.2-3B}}}                                                                                                                                                                          \\ 
\midrule
 \texttt{SMILES}                       & 43.57       & 40.40        & 40.15           & 56.13         & 48.62         & 45.77       & 0.00                              & -1.67                             \\
\texttt{MolT5}                         & 48.17       & 45.77        & 51.23           & 49.47         & 42.55         & 47.44       & 1.67                              & 0.00                              \\
\texttt{Llama-3}                     & 48.52       & 32.42        & 53.55           & 52.60         & 54.09         & 48.23       & 2.46                              & 0.80                              \\
\texttt{Llama-3.1}                   & 46.99       & 40.93        & 33.78           & 53.09         & 50.48         & 45.05       & -0.72                             & -2.39                             \\
\texttt{Ministral}                   & 49.65       & 47.18        & 50.76           & 47.79         & 50.82         & 49.24       & 3.47                              & 1.80                              \\
\texttt{Falcon3}                     & 59.43       & 44.96        & 39.81           & 52.51         & 42.38         & 47.82       & 2.05                              & 0.38                              \\
\texttt{ChemLLM}                      & 44.99       & 44.75        & 46.05           & 45.73         & 47.30         & 45.76       & -0.01                             & -1.68                             \\
\texttt{GPT-3.5}                      & 53.66       & 32.61        & 58.93           & 55.70         & 47.01         & 49.58       & 3.81                              & 2.14                              \\
\texttt{GPT-4o}                       & 50.12       & 42.20        & 38.42           & 59.38         & 54.33         & 48.89       & 3.12                              & 1.45                              \\ 
\midrule
\multicolumn{9}{c}{\textbf{\texttt{Llama-3.1-8B}}}                                                                                                                                                                          \\ 
\midrule
 \texttt{SMILES}                       & 38.10       & 37.56        & 35.30           & 52.70         & 44.89         & 41.71       & 0.00                              & -4.57                             \\
\texttt{MolT5}                         & 43.04       & 45.30        & 39.28           & 52.06         & 51.72         & 46.28       & 4.57                              & 0.00                              \\
\texttt{Llama-3}                     & 56.80       & 47.71        & 68.08           & 54.21         & 60.75         & 57.51       & 15.80                             & 11.23                             \\
\texttt{Llama-3.1}                   & 50.92       & 42.11        & 49.97           & 38.11         & 51.74         & 46.57       & 4.86                              & 0.29                              \\
\texttt{Ministral}                   & 59.73       & 50.18        & 70.01           & 51.65         & 55.59         & 57.43       & 15.72                             & 11.15                             \\
\texttt{Falcon3}                     & 57.92       & 51.11        & 53.22           & 42.38         & 47.57         & 50.44       & 8.73                              & 4.16                              \\
\texttt{ChemLLM}                      & 50.83       & 52.18        & 47.45           & 53.17         & 49.84         & 50.69       & 8.98                              & 4.41                              \\
\texttt{GPT-3.5}                      & 64.66       & 47.20        & 73.19           & 55.57         & 59.70         & 60.07       & 18.35                             & 13.79                             \\
\texttt{GPT-4o}                       & 56.64       & 53.37        & 55.28           & 52.60         & 61.65         & 55.91       & 14.20                             & 9.63                              \\ 
\midrule
\multicolumn{9}{c}{\textbf{\texttt{Llama-3.1-70B}}}                                                                                                                                                                         \\ 
\midrule
 \texttt{SMILES}                       & 49.21       & 39.40        & 52.12           & 45.98         & 47.73         & 46.89       & 0.00                              & -3.11                             \\
\texttt{MolT5}                         & 42.02       & 42.86        & 61.94           & 54.66         & 48.48         & 49.99       & 3.11                              & 0.00                              \\
\texttt{Llama-3}                     & 48.34       & 47.54        & 68.08           & 45.73         & 50.95         & 52.13       & 5.24                              & 2.14                              \\
\texttt{Llama-3.1}                   & 53.67       & 43.36        & 37.79           & 56.27         & 48.69         & 47.95       & 1.07                              & -2.04                             \\
\texttt{Ministral}                   & 59.85       & 60.62        & 68.31           & 51.22         & 48.28         & 57.66       & 10.77                             & 7.67                              \\
\texttt{Falcon3}                     & 54.12       & 40.02        & 60.75           & 43.24         & 53.02         & 50.23       & 3.34                              & 0.24                              \\
\texttt{ChemLLM}                      & 46.11       & 59.21        & 57.43           & 50.18         & 48.85         & 52.35       & 5.47                              & 2.36                              \\
\texttt{GPT-3.5}                      & 52.63       & 47.72        & 54.91           & 52.76         & 44.31         & 50.47       & 3.58                              & 0.47                              \\
\texttt{GPT-4o}                       & 53.69       & 62.98        & 50.23           & 56.67         & 53.07         & 55.33       & 8.44                              & 5.34                              \\ 
\bottomrule
\end{tabular}}
\caption{Full results (ROC-AUC \%) of model size experiments. $\boldsymbol{[Description]}$ indicates the hallucination source, whether generated by a specific model or from baseline descriptions.}
\label{table:full_results_model_size}
\end{table*}

\begin{table*}[ht]
\centering
\begin{tabular}{lcccccc}
\textbf{Model} & \textbf{HIV} & \textbf{BBBP} & \textbf{Clintox} & \textbf{SIDER} & \textbf{Tox21} & \textbf{Avg}  \\ 
\toprule
\texttt{Llama-3.2-1B}        & 10.26       & 13.06        & 10.54           & 10.43         & 9.74          & 10.81        \\
\texttt{Llama-3.2-3B}             & 10.16       & 10.06        & 8.41            & 9.59          & 10.61         & 9.77         \\
\texttt{Llama-3.1-8B}           & 7.15        & 8.04         & 7.24            & 7.57          & 7.11          & 7.42         \\
\texttt{Llama-3.1-70B}           & 8.37        & 6.82         & 7.31            & 8.60          & 6.62          & 7.54         \\
\bottomrule
\end{tabular}
\caption{HHM-2.1 scores for Llama models of different sizes.}
\label{tab:hallucination_model_size}
\end{table*}

\section{Full Results for Experiments on Generation Temperature}\label{sec:appendix_temperature}

Table \ref{table:full_results_temperature} presents the complete results of \texttt{Llama-3.1-8B} performance when using hallucinations generated at different temperatures. Additionally, the corresponding hallucination scores are shown in Table \ref{tab:hallucination_temperature}.

\begin{table*}[ht]
\centering

\begin{tabular}{lcccccc}
\textbf{Temperature} & \textbf{HIV} & \textbf{BBBP} & \textbf{Clintox} & \textbf{SIDER} & \textbf{Tox21} & \textbf{Avg}  \\ 
\toprule
0.1         & 61.82   & 64.85   & 49.77   & 54.13   & 54.93   & 57.10    \\ 

0.3         & 61.38   & 50.67   & 52.36   & 54.95   & 54.31   & 54.73    \\ 

0.5         & 60.03   & 48.11   & 69.48   & 50.10   & 57.38   & 57.02    \\ 

0.7         & 62.85   & 49.84   & 51.36   & 57.27   & 58.95   & 56.05    \\ 

0.9         & 61.34   & 45.12   & 56.20   & 48.69   & 59.66   & 54.20    \\
\bottomrule
\end{tabular}
\caption{Full results (ROC-AUC \%) of \texttt{Llama-3.1-8B} using hallucination generated by different generation temperatures.}\label{table:full_results_temperature}
\end{table*}

\begin{table*}[ht]
\centering
\begin{tabular}{lcccccc}
\textbf{Temperature} & \textbf{HIV}               & \textbf{BBBP}              & \textbf{Clintox}           & \textbf{SIDER}             & \textbf{Tox21}             & \textbf{Avg}                \\ 
\toprule
0.1                  & 8.02                       & 8.25                       & 7.52                       & 7.81                       & 8.07                       & 7.93                        \\
0.3                  & 8.57                       & 8.75                       & 6.87                       & 7.39                       & 8.05                       & 7.92                        \\
0.5                  & 8.35                       & 8.27                       & 7.83                       & 9.38                       & 5.96                       & 7.96                        \\
0.7                  & 8.10                       & 6.88                       & 7.37                       & 7.57                       & 8.71                       & 7.73                        \\
0.9                  & 6.61  & 6.85  & 6.42  & 7.96  & 6.27  & 6.82   \\
\bottomrule
\end{tabular}
\caption{The HHM-2.1 score for \texttt{Llama3.1-8B} generated hallucinations in different temperatures.}
\label{tab:hallucination_temperature}
\end{table*}

\section{Prompt Template for Annotating Hallucination in Types}\label{sec:appendix_prompt_annotation}

\begin{tcolorbox}[colback=skyblue!20, colframe=skyblue!80!black, width=0.48\textwidth, rounded corners]
\textit{System}: You are an expert annotator specializing in identifying hallucinations in molecular descriptions. Your task is to carefully analyze descriptions and categorize any hallucinations according to the provided taxonomy.\\
\textit{User}: $[SMILES]$ Your task is to identify
all the types of hallucinations present in the hallucinated description of the molecule. \\

Molecule: $[SMILES]$ \\
Hallucination: "$[HALLUCINATION]$"\\

Here is a description of each category:\\

$[CATEGORIES]$\\

Instructions:
\begin{enumerate}
\item Choose the category names that BEST matches the hallucination. Write these category names in decreasing order of importance.
\item Provide an explanation for why you chose this/these category name(s) and their order of importance.
\item Return json with molecule SMILES, category name(s) and explanation
\item If no category name matches, use "No hallucination"
\end{enumerate}
\label{colourbox:prompt}
\end{tcolorbox}

\end{document}